\documentclass{article}

\usepackage{natbib}
\usepackage{amsmath}
\usepackage{amsfonts}
\usepackage{graphicx}
\usepackage{physics}
\usepackage[scientific-notation=true]{siunitx}

\usepackage{microtype}
\usepackage{graphicx}
\usepackage{subcaption}
\usepackage{booktabs}
\usepackage{listings}
\usepackage{xcolor}
\usepackage{lipsum}
\usepackage{courier}

\usepackage{amsmath,amsfonts,bm}

\def\eqref#1{equation~\ref{#1}}

\def\1{\bm{1}}

\def\vtheta{{\bm{\theta}}}

\def\vh{{\bm{h}}}

\def\vx{{\bm{x}}}
\def\vy{{\bm{y}}}

\def\mX{{\bm{X}}}

\DeclareMathAlphabet{\mathsfit}{\encodingdefault}{\sfdefault}{m}{sl}
\SetMathAlphabet{\mathsfit}{bold}{\encodingdefault}{\sfdefault}{bx}{n}

\usepackage{hyperref}

\usepackage[preprint]{neurips_2020}

\DeclareMathOperator*{\expect}{\mathop{\mathbb{E}}}

\title{Evaluating Prediction-Time Batch Normalization for Robustness under Covariate Shift}

\author{
  Zachary Nado\\
  Google Research, Brain Team\\
  \texttt{znado@google.com}
  \And
  Shreyas Padhy\thanks{AI Resident}\\
  Google Research, Brain Team
  \And
  D. Sculley\\
  Google Research, Brain Team
  \And
  Alexander D’Amour\\
  Google Research, Brain Team
  \And
  Balaji Lakshminarayanan\\
  Google Research, Brain Team
  \And
  Jasper Snoek\\
  Google Research, Brain Team
}

\begin{document}

\maketitle

\begin{abstract}
Covariate shift has been shown to sharply degrade both predictive accuracy and the calibration of uncertainty estimates for deep learning models. This is worrying, because covariate shift is prevalent in a wide range of real world deployment settings. However, in this paper, we note that frequently there exists the potential to access small unlabeled batches of the shifted data just before prediction time.  This interesting observation enables a simple but surprisingly effective method which we call prediction-time batch normalization, which significantly improves model accuracy and calibration under covariate shift.  Using this one line code change, we achieve state-of-the-art on recent covariate shift benchmarks and an mCE of 60.28\% on the challenging ImageNet-C dataset; to our knowledge, this is the best result for any model that does not incorporate additional data augmentation or modification of the training pipeline. We show that prediction-time batch normalization provides complementary benefits to existing state-of-the-art approaches for improving robustness (e.g. deep ensembles) and combining the two further improves performance. Our findings are supported by detailed measurements of the effect of this strategy on model behavior across rigorous ablations on various dataset modalities. However, the method has mixed results when used alongside pre-training, and does not seem to perform as well under more natural types of dataset shift, and is therefore worthy of additional study. We include links to the data in our figures to improve reproducibility, including a Python notebooks that can be run to easily modify our analysis at \href{https://colab.research.google.com/drive/11N0wDZnMQQuLrRwRoumDCrhSaIhkqjof}{this url}.
\end{abstract}

\section{Introduction}
Covariate shift is one of the key problems facing modern machine learning. 
Informally defined as situations in which training data
differs from the data seen at final prediction time, covariate shift breaks the traditional {i.i.d.} assumptions
used to underpin supervised machine learning \citep{vapnik95}.  For deep models in particular,
covariate shift has been shown to not only cause incorrect predictions, but to do so with
disproportionately high levels of confidence \citep{ovadia2019}.  This is potentially worrying,
because covariate shift often occurs in practical settings such as real-world deployment
of ML systems \citep{mcmahan2013ad}, for reasons that may include non-stationarity over time or differences between
local and global distributions.  Furthermore, traditional training-time methods for covariate shift
correction may be impractical in such settings.

In this paper, we make the observation that modern machine learning systems utilize
{\em batching} at prediction time for reasons of computational efficiency, especially with hardware such as GPUs and TPUs that amortize cost well across batches of hundreds or thousands of examples
\citep{jouppi2017datacenter}.
In this setting, examples are batched together -- often from a stream of data that may include
many thousands of examples per second -- creating a brief window of opportunity to examine
the data characteristics of this small, unlabeled batch of prediction-time data.  When deployed
models are replicated globally, as in the setting described by \citet{mcmahan2013ad}, each local model may see different distributions that reflect local data characteristics.  Application areas that can fall into this setting include large scale systems for image recognition and ad click through predictions \citep{mcmahan2013ad}.

Because the window of opportunity to examine the prediction-time batch is brief, often measured on the order of milliseconds before predictions must be made, any practical correction technique
in this setting must be computationally efficient and operate without resorting to methods
that re-train models.  

We propose one such method, which we call prediction-time
batch normalization, as an extension of the widely adopted (training time) batch normalization method proposed by \citep{batchnorm}.
We motivate the development of prediction-time batch norm by empirically analyzing the activations within hidden layers of deep learning models to understand the cause of mistakenly overconfident behavior under covariate shift.  Experimentally, we demonstrate that prediction-time batch norm provides a simple, computationally
efficient method that yields surprisingly effective results on the distributional shift benchmark of~\citet{ovadia2019}.  Furthermore, prediction-time batch norm achieves an mCE of 60.28\% on the challenging ImageNet-C benchmark, which is the best result to our knowledge for a model that does not incorporate additional data augmentation.

\textbf{Contributions} In this paper, we formalize a perhaps underappreciated prediction setting, which we call the {\em prediction-time batch setting}.  We then propose a simple method, prediction-time BN, for using this information to effectively correct for covariate shift, motivating this approach by analyzing properties
of deep models and their internal activations under covariate shift.
We demonstrate that this works very effectively in practice on multiple modalities.
We also carefully analyze the model performance, teasing apart several factors to help understand why this method works well, identifying key factors via ablation studies.
Finally, we explore the limits of the method under more natural types of dataset shift and examine its potential failure modes, notably its lackluster performance when combined with pre-training, which indicates that the method is worthy of additional study.
Together the results in this paper lay out an interesting and highly practical methodology which could be used, with caveats, to correct for covariate shift in real-world deployment settings.

\section{Setup}
\subsection{The Prediction-Time Batch Setting}
Here, we formalize our prediction setting.
We observe feature-label pairs $\{(\vx_i, y_i)\}_{i=1}^N$ drawn i.i.d. from some training distribution $p(\vx, y)$, and wish to predict the labels of unlabeled test examples $\{\vx_j\}_{j=1}^T$.
We assume that the test examples are drawn i.i.d. from a potentially distinct, unknown target distribution $q(\vx)$, with an accompanying conditional label distribution $q(y \mid \vx)$.

Unlike the standard supervised learning setting, we assume that predictions can be made in batches.
Specifically, at prediction time, we obtain batches of $t < T$ examples, $\mX^{(b)} = (\vx^{(b)}_i)_{i=1}^t$, and make predictions for these examples simultaneously, $\hat \vy^{(b)} = (\hat y^{(b)}_1, \ldots, \hat y^{(b)}_t) = f_\vtheta(\mX^{(b)})$.
We express our goal as minimizing a predictive risk that is evaluated batch-wise.
Specifically, let $\ell(\vy^{(b)}, f_\vtheta(\mX^{(b)})) := \sum_{i=1}^t \ell(y^{(b)}_i, \hat y^{(b)}_i)$ be a batch-wise loss function that decomposes additively across the points in a test batch. 
Our goal is to minimize the expected loss, or risk, over i.i.d. test batches $(\mX, \vy)$ drawn from the product distribution $q(\vx, y)^t := \prod_{i=1}^t q(\vx_i, y_i)$:
\begin{align}
\label{eq:batch pred risk}
\min_\vtheta \expect_{(\mX, \vy) \sim q(\vx, y)^t}\left[\ell(\vy, f_\vtheta(\mX))\right].
\end{align}
Conveniently, because the loss function $\ell$ decomposes linearly, the empirical analogue of this risk can be computed in the standard way, as the mean loss across examples $\ell(y_i, \hat y_i)$. 
Thus, empirical evaluation in this setting can be done using standard pipelines.\footnote{The within-batch dependence introduced by simultaneous predictions $f_\vtheta(\mX^{(b)})$ will generally alter the concentration properties of this empirical mean, but this effect is modest if the batch size is small relative to the size of the test set.
We conjecture that, in the worst case, generalization bounds scale in the number of batches rather than the number of examples.
}

\subsection{Reliable Uncertainty Quantification Under Covariate Shift}

Our goal is to produce a batch-wise prediction function $f_\vtheta(\mX)$ that appropriately expresses uncertainty when the training and test distributions differ, $q(\vx, y) \neq p(\vx, y)$.
In particular, we focus on the covariate shift setting \cite{Shimodaira2000,Quionero-Candela2009}, where the marginal distributions of features are different, $p(\vx) \neq q(\vx)$, but the conditional label distributions are the same, $p(y \mid \vx) = q(y \mid \vx)$.
In this setting, an effective method should return predictions that are less confident for inputs $\vx$ that are rare under the training distribution $p(\vx)$ or are outside of its support.

Formally, we quantify the quality of the uncertainty of a predictive distribution using two measures, \emph{calibration error} and \emph{Brier score}.
\textbf{Expected Calibration Error} (ECE)~\citep{guo2017calibration} is the difference between the confidence and accuracy of a model, binned by confidence. We define confidence as the max predicted class probability for a given example. Let $B_i$ be the elements in each confidence bin, then
$\text{ECE} = \sum_{i}{\frac{B_i}{N}|\operatorname{acc}\left(B_i\right) - \operatorname{conf}\left(B_i\right)|}$,
where $\operatorname{acc}\left(B_i\right)$ is the accuracy of elements in $B_i$ and $\operatorname{conf}\left(B_i\right)$ is the confidence of elements in $B_i$.\footnote{Note that this metric can be sensitive to the number of bins used, and can yield very low scores even when the model is predicting poorly (such as when the model outputs the uniform distribution and the model achieves chance accuracy.) We use 10 bins for CIFAR-10-C experiments and 30 bins for ImageNet-C.}
\textbf{Brier Score}~\citep{brier1950verification} is defined as the squared distance between a model output distribution and the one-hot target labels. It is a proper scoring rule \cite{gneiting2007strictly}, and as such decreases to zero monotonically as the predictive distribution approaches the true underlying distribution.

\section{Related work}\label{sec:related_work}
There is a large body of work addressing covariate shift issues in the literature on domain adaptation.
\citet{wilson2018survey} survey different methods in unsupervised domain adaptation.
These include methods that learn mappings between domains \citep{fernando2013unsupervised, sener2016learning}, match means and covariances across feature vectors \citep{sun2017correlation}, or match moments of the distributions directly \citep{peng2019moment} or through kernel embeddings \citep{long2015learning, gong2012geodesic}.
Importance weighting methods are also a common approach,
where training examples are reweighted to minimize an estimate of the predictive risk in the target domain~\citep{sugiyama2007}.
A common theme in these approaches is that they assume access to a set of unlabeled samples from the test set at training time \citep{sun2019test}, whereas our focus is to apply an intervention at test time, after a model has been trained.
In addition, these methods can become brittle when the training and test distributions are highly distinct~\citep{johansson2019support}.

There has been work in the domain adaptation literature that specifically focuses on using batch normalization as a means of mapping between domains, and \citet{li2016revisiting} make the claim that the batch norm statistics in deep networks learn domain-specific knowledge. They propose AdaBN, which calculates domain-specific batch norm statistics using the entirety of the target domain at test-time (or an exponential moving average (EMA) in practice). Subsequent work has expanded upon this idea, however many of the methods proposed require access to the target domain data during training time. AutoDIAL \citep{cariucci2017autodial} mixes data from the source and target domains during training time before passing them through the batch norm layers, whereas TransNorm \citep{wang2019transferable} uses source and target domain batch norm statistics during training in an end-to-end fashion to improve transferability across domains. To our knowledge, AdaBN was the only previous normalization strategy used in domain adaptation literature that applies a test-time correction, and they require access to the entirety of the target domain during evaluation. A similar concurrent work is \citep{schneider2020improving}, which further demonstrated the usefulness of using batch normalization statistics for improving robustness under covariate shift.

While \citet{guo2017calibration} observed that models using traditional batch norm typically have worse calibration on the test set, to our knowledge no one has applied normalization strategies for correcting miscalibration under covariate shift. Pre-existing methods that improve calibration performance on the image datasets we consider, such as mCE on ImageNet-C, either involve extensive data augmentation strategies ~\cite{hendrycks2019augmix,lopes2019improving,chun2019empirical,lee2020compounding} or require substantial pre-training \citep{xie2019self}.

Normalization methods have also been used in generative modelling tasks, where modifying normalization statistics is useful for generating distinct images of the same object \citep{miyato2018spectral, de2017modulating, dumoulin2016learned}. \citet{li2018twin} specifically make use of batch normalization for domain adaptation. The prediction-time batch setting has also been explored for deep generative models, particularly in the context of addressing their failure modes for OOD detection as reported by \citet{nalisnick2018deep,choi2018generative}. \citet{nalisnick2019ood} propose a typicality test for generative models that performs OOD detection using a batch of inputs. \citet{song2019unsupervised} use prediction-time batch normalization in deep generative models and show that it improves OOD detection. This is a closely related work, but is complementary to our method, as we focus on discriminative models and covariate shift rather than OOD detection.

\begin{figure}[!h]
    \centering
      \includegraphics[width=0.85\linewidth]{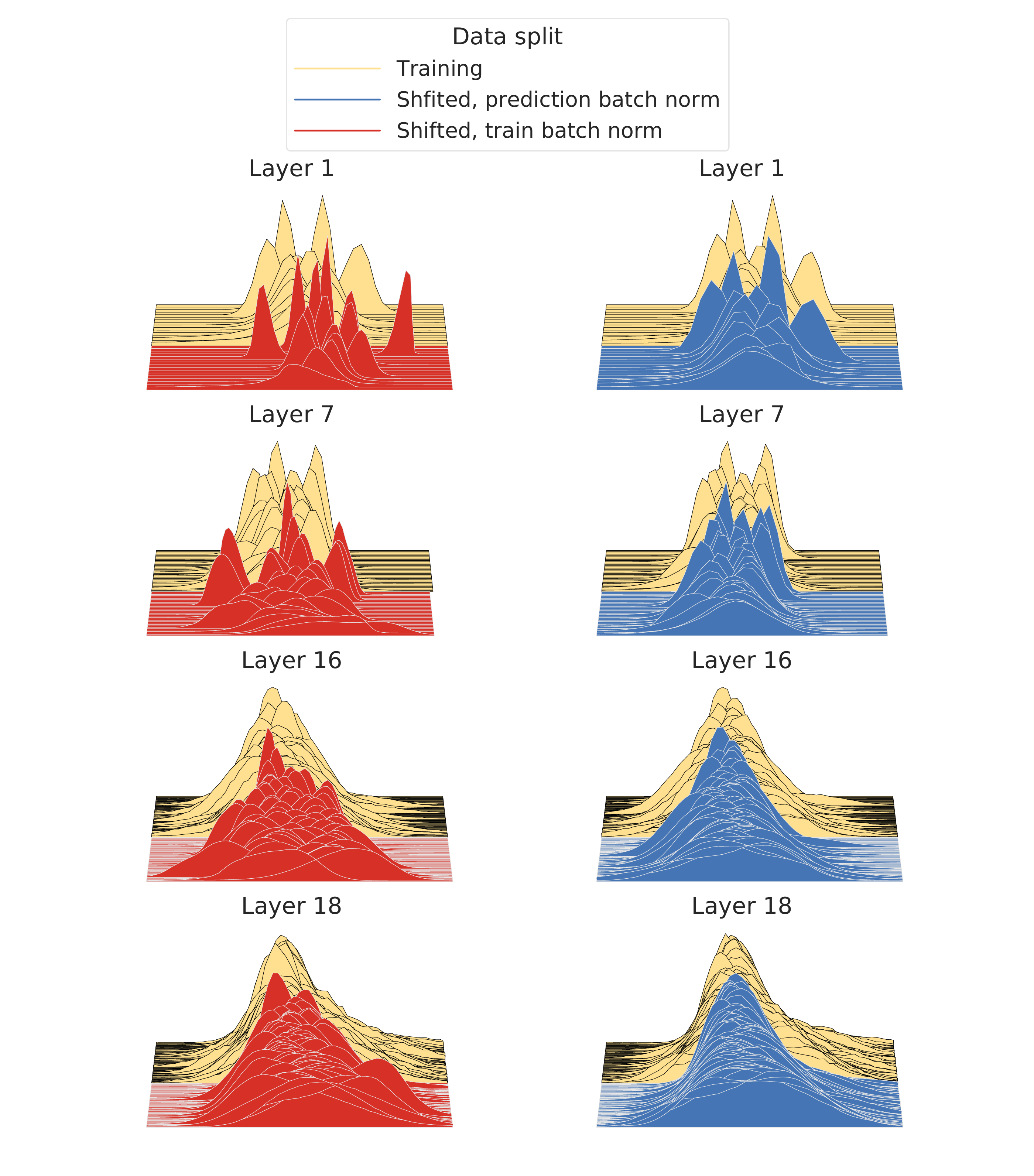}
    \caption{\textbf{Empirical distributions for the output of selected normalization layers in Resnet-20 on CIFAR10 and CIFAR10-C.} Activations are averaged over spatial dimensions, resulting in one distribution per output channel. The activations are recorded immediately after the batch normalization layer, before the non-linearity of each layer. The blue and red curves are aggregated across all shifted examples, while the yellow is across all training examples. We can clearly see that prediction-time BN is much more effective at aligning the shifted activations with the training distribution support and shape. These layers were picked as representative examples of activations of all normalization layers in the model, we encourage the reader to check Figures~\ref{fig:cifar10_bn_activation_dists_ema_all}, ~\ref{fig:cifar10_bn_activation_dists_test_batch_all} for all layers.}
    \label{fig:cifar10_bn_activation_dists_select}
\end{figure}

\clearpage\newpage

\section{Prediction-Time Batch Normalization}
We propose a simple protocol to mitigate the effects of covariate shift at prediction time in a batch-prediction setting: use batch normalization (BN) with statistics recalculated on the batch at prediction time.

Specifically, batch normalization~\cite{batchnorm} normalizes the pre-activations, denoted by $\hat{x}_i^{\left(c\right)}$ for channel $c$, of a layer:
\begin{align}\label{eq:batchnorm}
    \hat{x}_i^{\left(c\right)} = \frac{x_i^{\left(c\right)} - \mu^{\left(c\right)}}{\sqrt{\sigma^{\left(c\right)^2} + \epsilon}};\;\;\;\\
    \mu^{\left(c\right)} = \frac{1}{NHW}\sum_{n,h,w}{x_{nhw}},\;\;\;
    \sigma^{\left(c\right)^2} = \frac{1}{NHW}&\sum_{n,h,w}{\left(x_{nhw} - \mu^{\left(c\right)}\right)^2}
\end{align}
where $N$, $H$, and $W$ are the number of examples in a batch and their dimensionality (height and width).
In standard practice, the batch normalization statistics $\mu^{(c)}$ and $\sigma^{(c)^2}$ are frozen to particular values after training time that are used to compute predictions, an approach we refer to as \textbf{train BN}.
On the other hand, our strategy, \textbf{prediction-time BN}, recomputes these statistics for each test batch.
While prediction-time batch normalization has been explored in the literature on domain adaption and the literature on OOD detection using deep generative models, we are not aware of any prior work investigating its usefulness for robust deep learning under covariate shift (see Section~\ref{sec:related_work} for a discussion).

In our experiments, we find that prediction-time BN is surprisingly effective for improving uncertainty quantification in deep networks (Section~\ref{sec:results}).
To move toward an understanding of why this is the case, we offer two observations here.
We then perform a more thorough exploration of this method with ablation studies in Section~\ref{sec:ablations}.

\paragraph{Prediction-Time Batch Normalization Repairs Mismatched Supports}

\begin{figure}[!t]
    \centering
      \includegraphics[width=0.9\linewidth]{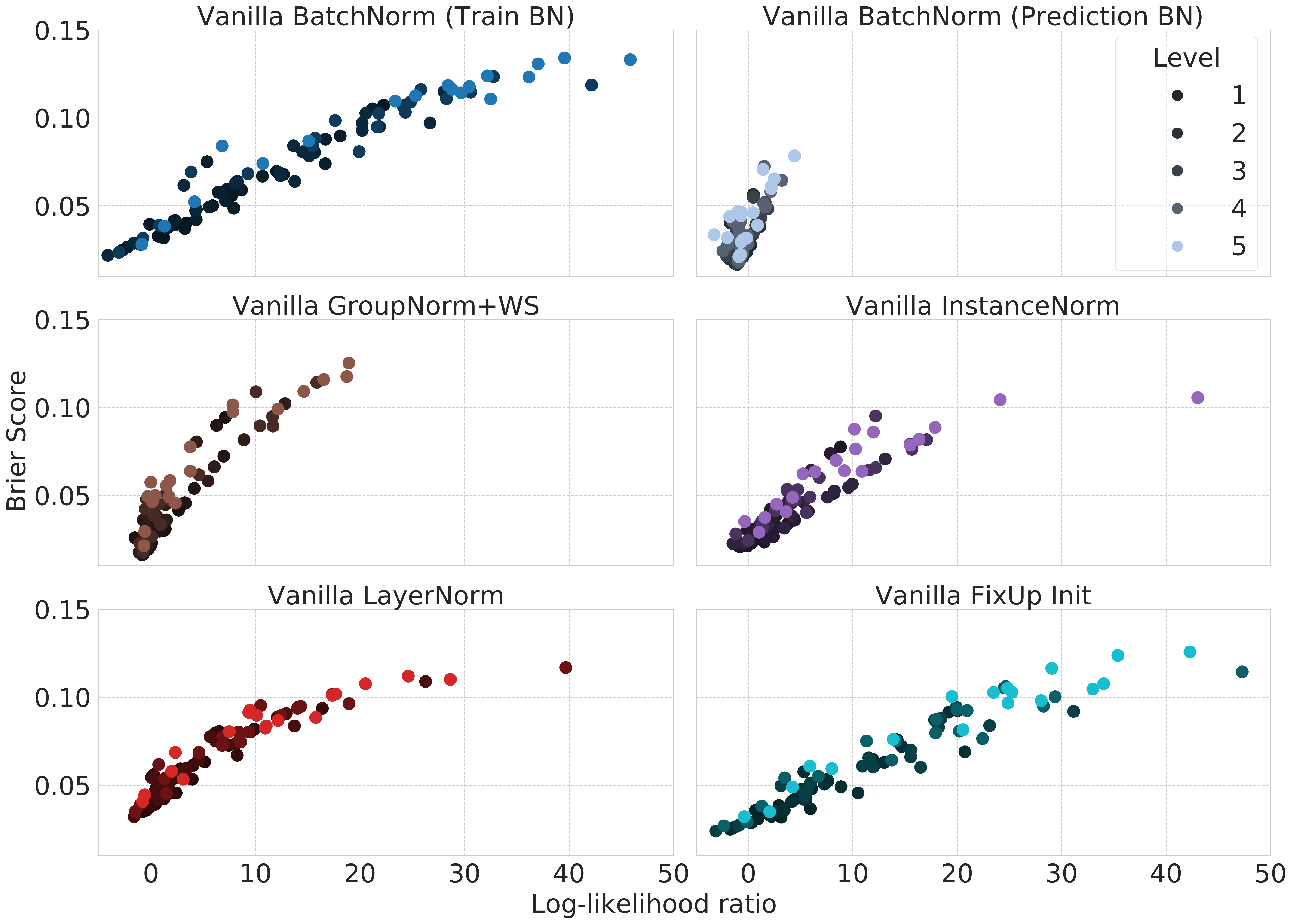}
    \caption{\textbf{Brier scores of predictions become higher when the activations from the training and test sets occur in increasingly distinct regions.}
    Here, we summarize how the distributions of penultimate hidden layer activations $\vh = g\left(\vx \right)$ on shifted test sets compare to their distributions on the training set, under a number of different normalization schemes.
    Each point represents a type of shift, where the color indicates the intensity of the shift applied. 
    On the horizontal axis, we plot a measure of the discrepancy between the training and test distributions of activations, $p(\vh)$ and $q(\vh)$, respectively by approximating $KL(p(\vh) \| q(\vh)) \approx T^{-1}\sum_{i=1}^T \ln\frac{\hat q(\vh_i)}{\hat p(\vh_i)}$, where the summation is taken over test instances, and $\hat p$ and $\hat q$ are multivariate normal densities whose means and variances match $p$ and $q$, respectively.
    We use KL divergence because it is particularly sensitive to cases where test activations lie outside of the effective support of training activations.
    On the vertical axes are the Brier scores for each shifted example, averaged within each split.
    In addition to the overall trend of increasing discrepancy leading to decreasing performance, we also see that higher shift intensities tend to have higher support mismatch. 
    Intuitively, more shift in the inputs should lead to more shift in the activation values.
    We can clearly see that only when using prediction-time BN to compute $g\left(\vx\right)$ are the activations more closely aligned and the Brier Scores more consistently lower.
    See Figure~\ref{fig:cifar10_likelihood_ratio_scatter_all} for similar trends in accuracy.}
    \label{fig:cifar10_likelihood_ratio_scatter}
\end{figure}

As $q\left(\vx\right)$ shifts, the internal activations of a deep model $f_\theta(\vx)$ can move outside the ranges encountered during training, as seen in the left column of Figure~\ref{fig:cifar10_bn_activation_dists_select}.
When this happens, the model layers receive inputs outside of the domain they were trained on, and we can no longer expect well-defined model behavior such as accurate or well-calibrated predictions.
The top left panel of Figure~\ref{fig:cifar10_likelihood_ratio_scatter} demonstrates this behavior, with shifted examples inducing worsening calibrations as measured using Brier Score.
Overall there is a trend of decreasing calibration as the distance between test and train activations increases (see Figure~\ref{fig:cifar10_likelihood_ratio_scatter_all} for a similar trend in accuracy).

In Figure~\ref{fig:cifar10_likelihood_ratio_scatter}, we visualize the discrepancy between the supports of the empirical training and test activation distributions for the penultimate hidden layer of Resnet-20 on CIFAR10-C.
We see that the prediction-time BN correction is effective at bringing the activation distribution supports into alignment (clustered around 0 on the horizontal axis) relative to both train BN and other normalization schemes.

\paragraph{Prediction-Time Batch Normalization Maps Activations to Regions of Uncertainty}
Aligning the supports of activation distributions is not sufficient to ensure well-calibrated predictive distributions. 
For example, it would be plausible that a normalization scheme could spuriously map the activations of out-of-support test instances to regions in the activation space that induce highly confident predictions.
We find that prediction-time BN avoids this potential failure mode.
As Figure~\ref{fig:cifar10_likelihood_ratio_scatter} suggests, prediction-time BN seems to map out-of-support activations to regions in the training activation support that induce uncertain predictions, resulting in consistently lower Brier scores when the normalization is applied.
This pattern is confirmed in Figure~\ref{fig:imagenet_reliability}.

\clearpage\newpage

\section{Performance Under Covariate Shift}\label{sec:results}

To evaluate model behavior under covariate shift, we follow the benchmarking methodology
of \citep{ovadia2019} and evaluate our models on CIFAR-10-C, ImageNet-C ~\citep{hendrycks2019benchmarking}, and corrupted versions of the Criteo Display Advertising Challenge\footnote{\tiny \url{https://www.kaggle.com/c/criteo-display-ad-challenge}} (see Appendix~\ref{appendix:datasets} for more details). We include two data modalities in order to confirm that our strategy is effective on more than just image problems. The two image datasets contain versions of the original test dataset, with 19 different types of corruptions applied with 5 levels of shift intensity each. We refer to one of these 95 shifted versions of the test set as a \textit{split}.

\begin{figure}[!h]
    \centering
    \begin{subfigure}{0.99\textwidth}
        \centering
          \includegraphics[width=0.9\linewidth]{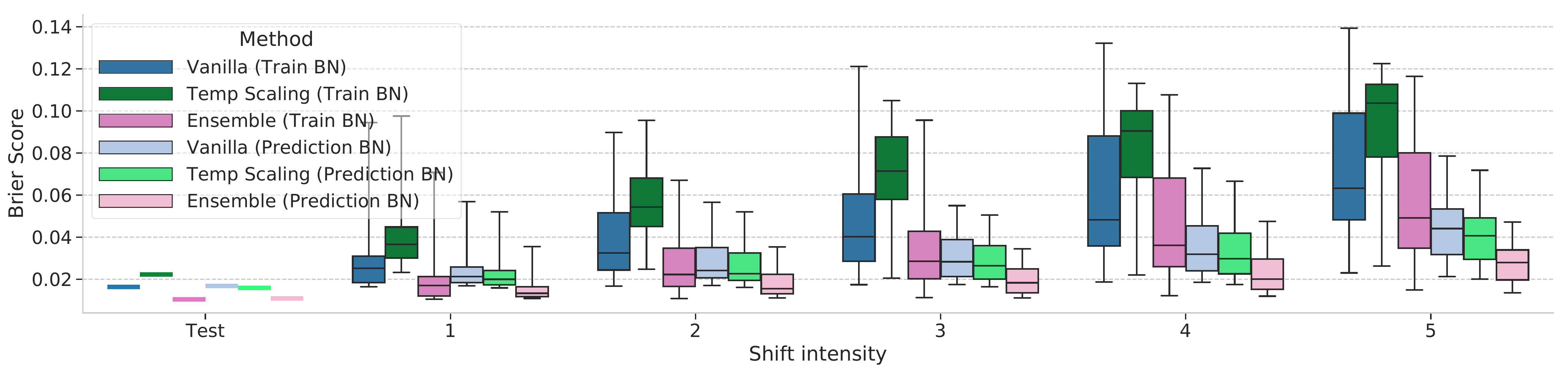}
        \caption{\textbf{Calibration under covariate shift for CIFAR-10-C with a prediction batch size of 500 (lower Brier Score is better).} Ensembles with prediction batch norm appear minimally affected by the level of shift. See Figure~\ref{fig:cifar10_methods_comparison_skew_all} for other metrics. The data and notebook for this plot can be found at \url{https://tensorboard.dev/experiment/IwvHAvuxTZK00Wp76rJwqw/}.}
    \end{subfigure}
    \vspace{2em}
    \begin{subfigure}{0.99\textwidth}
        \centering
          \includegraphics[width=0.9\linewidth]{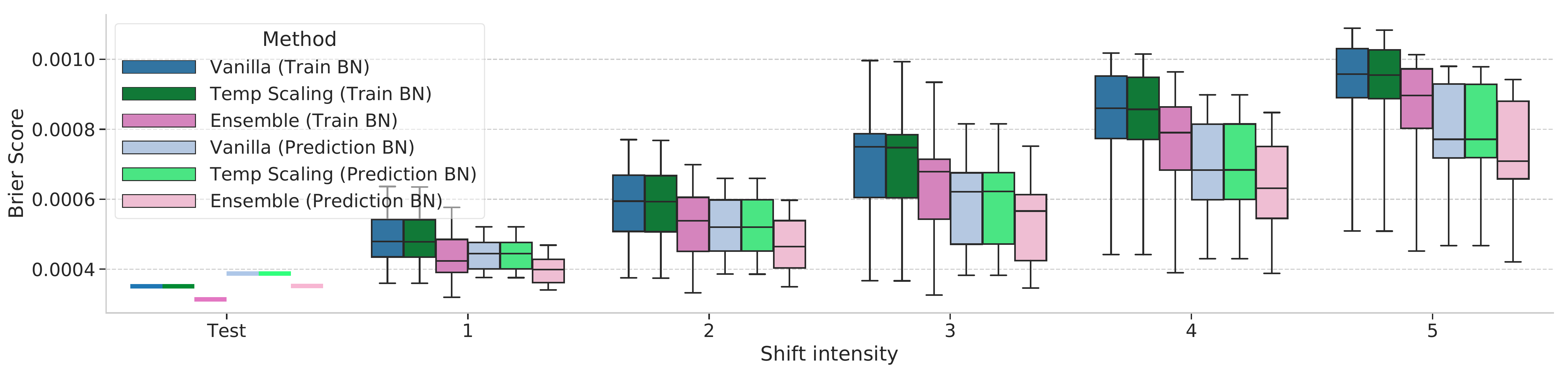}
        \caption{\textbf{Calibration under covariate shift for ImageNet-C with a prediction batch size of 100.} See Figure~\ref{fig:imagenet_methods_comparison_skew_all} for other metrics. The data and notebook for this plot can be found at \url{https://tensorboard.dev/experiment/FRbuxfG5SkaFPQQH4OcpYw/}.}
    \end{subfigure}
    \vspace{2em}
    \begin{subfigure}{0.99\textwidth}
        \centering
          \includegraphics[width=0.9\linewidth]{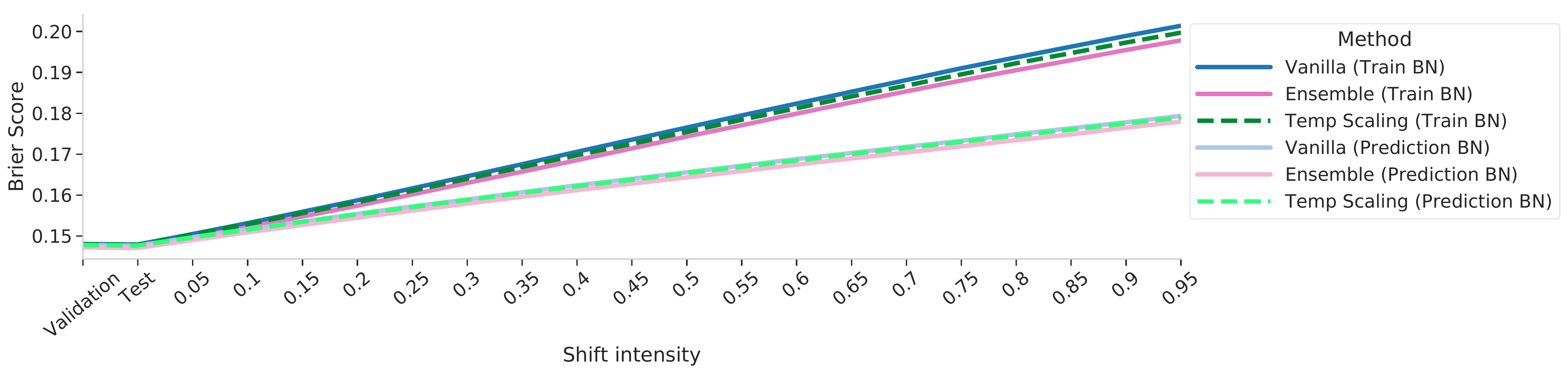}
        \caption{\textbf{Calibration under covariate shift for Criteo with a prediction batch size of 500.} See Figure~\ref{fig:criteo_basic_method_compare_all} for AUC performance. The data and notebook for this plot can be found at \url{https://tensorboard.dev/experiment/dNxyMRncRgSzozlD1m94Og/}.}
    \end{subfigure}
    \caption{Calibration across CIFAR10-C, ImageNet-C, and Criteo across increasing levels of dataset shift. The box plots show the median, quartiles, minimum, and maximum performance per method.}
    \label{fig:basic_method_compares}
\end{figure}

\clearpage\newpage

\subsection{Methods}
In our empirical analyses, we consider methods that are test-time only modifications of neural networks.  Our method should be applicable to any model with batch normalization. Many sophisticated Bayesian techniques, such as those explored in ~\citet{ovadia2019}, also require modifications at training time, which is not the setting we consider.
\textbf{Vanilla} refers to a neural network with a sigmoid or softmax final layer to produce a probability distribution over class labels. For more details on model architectures, see Appendix~\ref{appendix:models}.

\textbf{Ensemble}~\cite{deepensembles} is the same base model trained $M$ times, each with a different random seed, and the individual model predictions averaged post-softmax. We use $M = 10$ for all experiments.

\textbf{Temperature Scaling}~\citep{guo2017calibration,platt99} involves post-hoc tuning of a softmax temperature parameter on the validation data.

\textbf{Batch Normalization} ~\citep{batchnorm} traditionally normalizes along all but the last dimension. Batch normalization also computes an exponential moving average (EMA) of the batch means and variances throughout training, which is then used to normalize activations at test time. This has the benefit of making test-time predictions independent of the other elements in the test batch, and removes the effect of the test-time batch size on the normalization.

\textbf{Instance Normalization} ~\citep{instancenorm} is a version of Batch Norm which normalizes only along the spatial dimensions per batch element.

\textbf{Layer Normalization} ~\citep{layernorm} was originally proposed for recurrent neural networks, and normalizes along all but the batch dimension.

\textbf{Group Normalization} ~\citep{groupnorm} is similar to layer norm, but instead of normalizing over all channels at once it splits them into subgroups to compute statistics. We use two groups of channels for normalization. Weight Standardization~\citep{weightstd} is also used in all experiments with Group Norm as is common to improve model performance.

\subsection{Combining Normalization with Other Methods}

In addition to our vanilla model, prediction-time BN is complementary to other methods, as seen in Figure~\ref{fig:basic_method_compares}. Summarizing our ImageNet-C results, we achieve an mCE of 60.28\% with prediction-time BN using our Vanilla (Resnet-50) model.
For Criteo, as expected the AUC and Brier Score degrade with increasing levels of shift for both batch norm strategies (see Figure~\ref{fig:criteo_basic_method_compare_all}), but prediction-time BN relaxes this trend.

\subsection{Prediction Batch Dependence}
\begin{figure}[!h]
    \centering
      \includegraphics[width=0.7\linewidth]{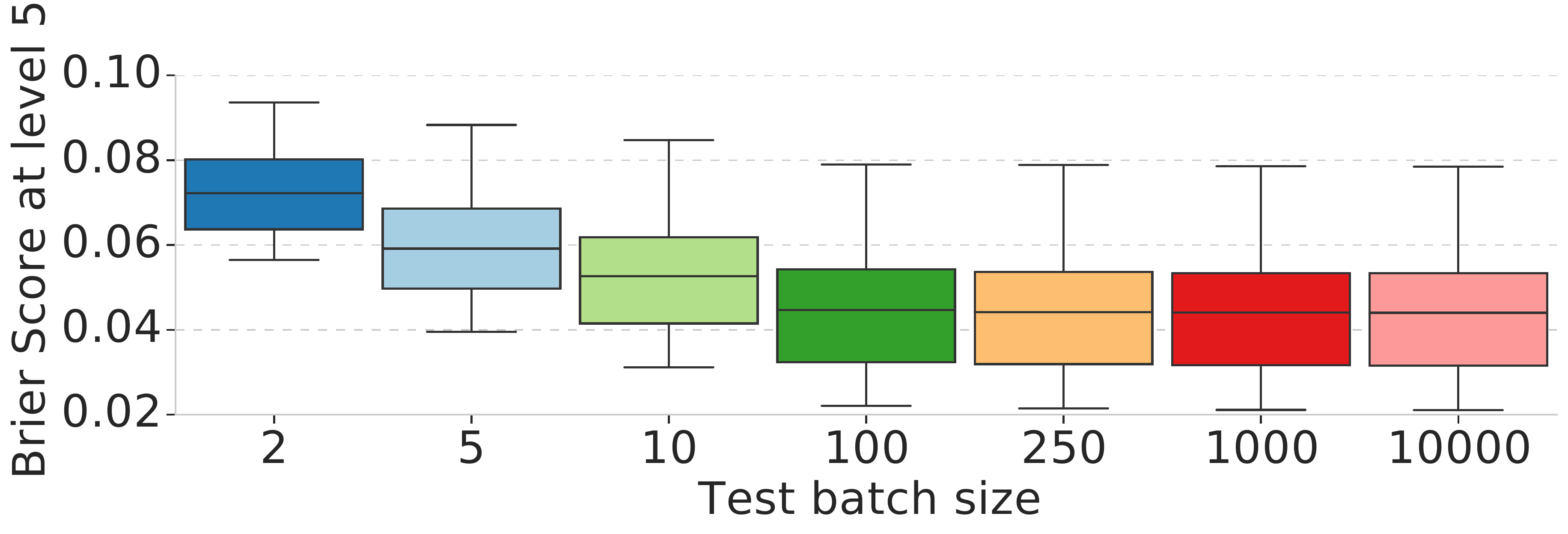}
    \caption{\textbf{CIFAR-10-C Brier Score at shift level 5 for different prediction batch sizes.} We see that relatively small batch sizes are required to effectively correct for covariate shift, with only marginal improvements after a 100 examples.  See Figure~\ref{fig:cifar_test_bs_all} for how other metrics perform across prediction batch sizes, and Figure~\ref{fig:imagenet_test_bs_all} for a similar trend on ImageNet-C.}
    \label{fig:cifar_test_bs_single_brier_score}
\end{figure}

One potential concern with prediction-time BN is that predictions now depend on other examples and the prediction batch size. However, as seen in Figure~\ref{fig:cifar_test_bs_single_brier_score} we evaluate performance across a range of prediction batch sizes and see we achieve strong performance with a batch of just 100 examples, with larger batch sizes giving marginal improvements.

\begin{figure}[!h]
    \centering
      \includegraphics[width=0.7\linewidth]{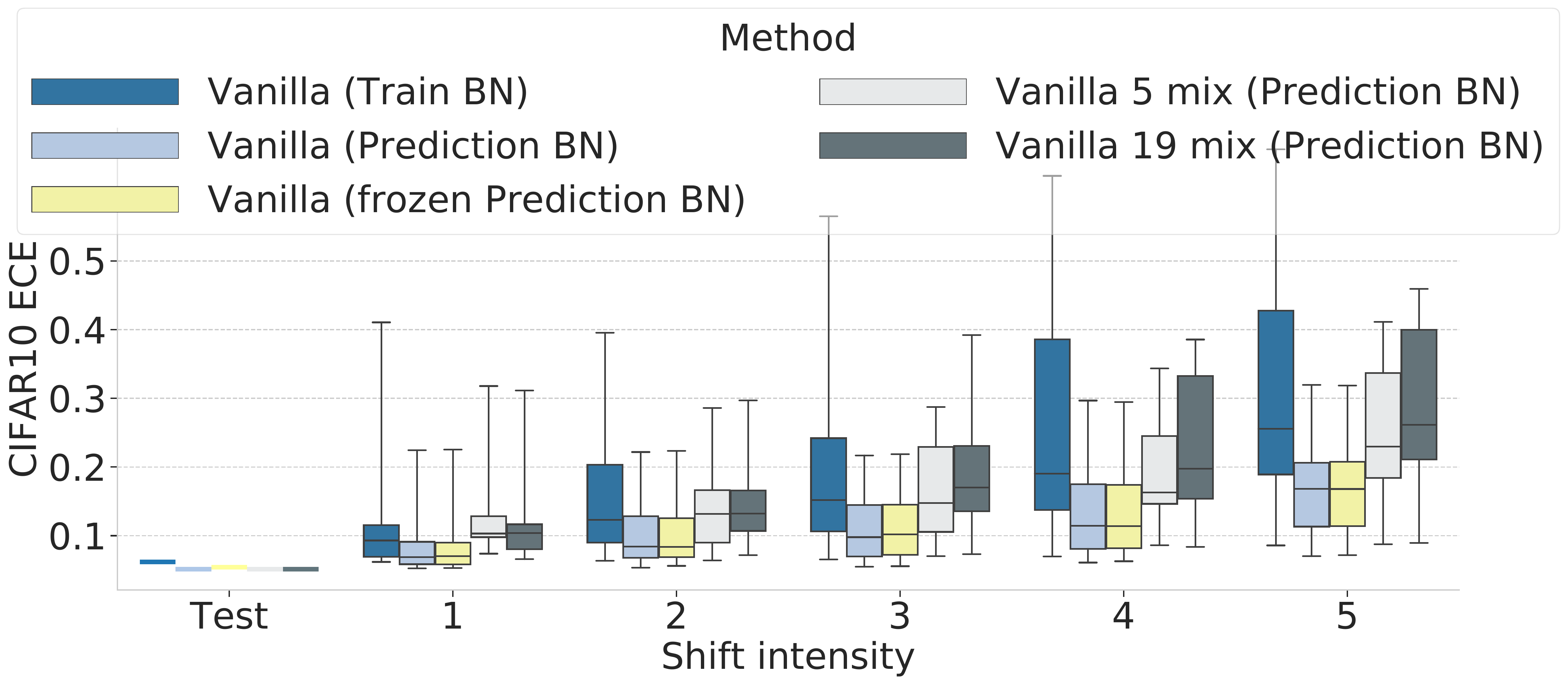}
      \includegraphics[width=0.7\linewidth]{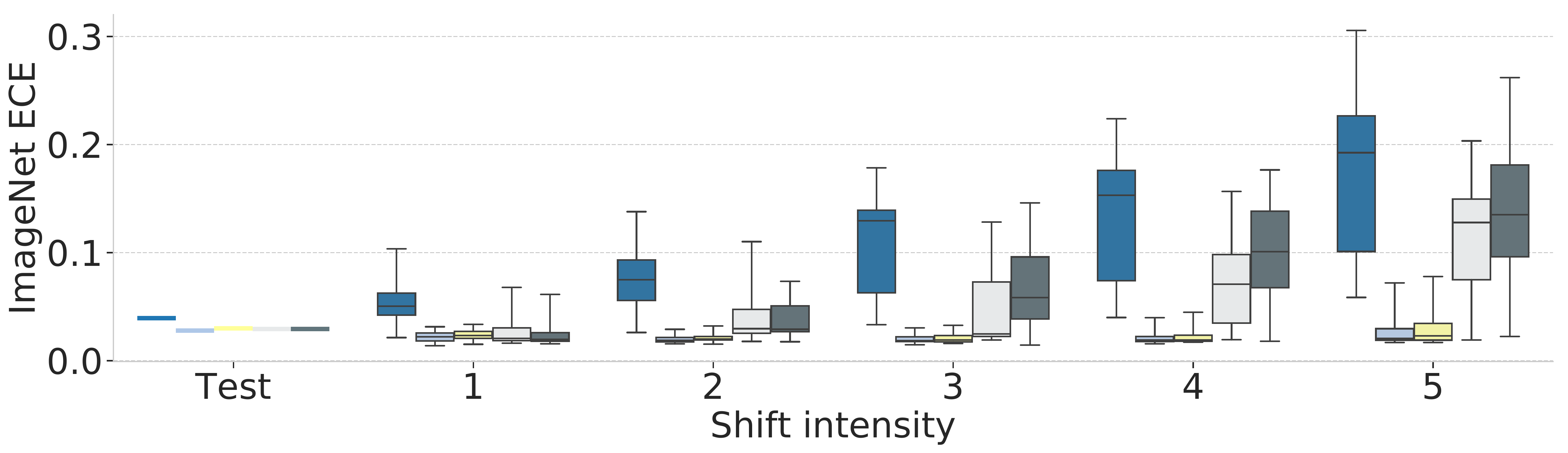}
    \caption{\textbf{Calibration on CIFAR-10-C (top) and ImageNet-C (bottom), both with a prediction batch size of 500.} Here we clearly see that just having access to a single batch from each split (frozen Prediction BN) is sufficient to get substantial performance improvements. Also, while prediction-time BN is sensitive to multiple simultaneous types of covariate shift, it still outperforms train BN. See Figures~\ref{fig:cifar_bn_stat_method_all}, \ref{fig:imagenet_bn_stat_method_all} for the accuracy and Brier score performance.}
    \label{fig:bn_stat_method_ece}
\end{figure}

We also investigate whether the performance gains are from having the statistics of the exact batch we are predicting on, or just from having statistics that are relevant to the prediction distribution's shift. In Figure~\ref{fig:bn_stat_method_ece} the statistics from the first batch of each shifted split are stored and reused for all subsequent batches of that split. We only see a marginal performance decrease, meaning we can remove our prediction batch dependence while retaining the performance improvements. Additionally, in Figure~\ref{fig:bn_stat_method_ece}, we test calibration under multiple corruption types and levels at once, and see that even with up to 19 different types of simultaneous shift, prediction-time BN outperforms train BN.

\subsection{Limitations}
\begin{figure*}[!h]
    \centering
      \includegraphics[width=0.95\linewidth]{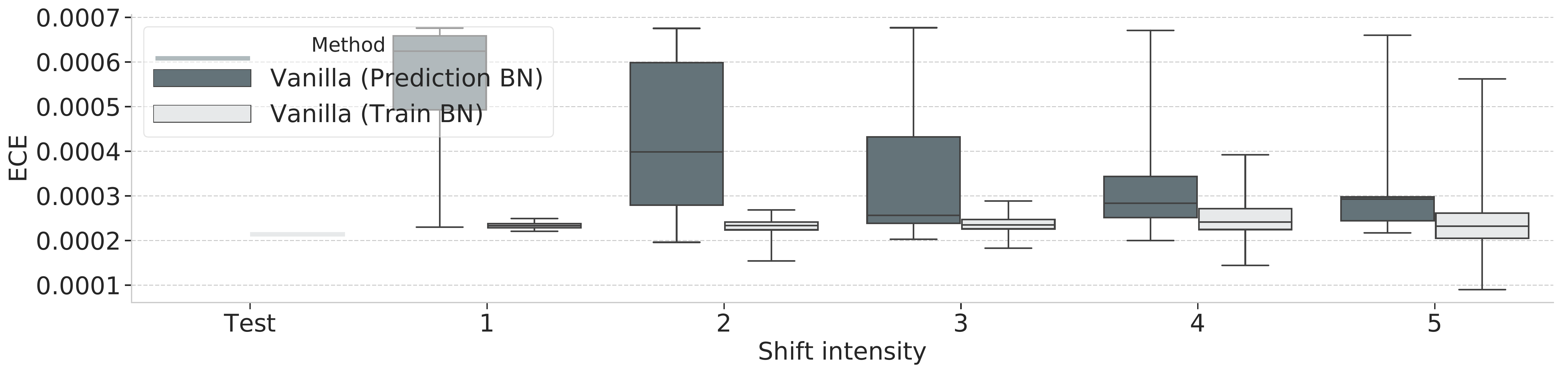}
      \includegraphics[width=0.95\linewidth]{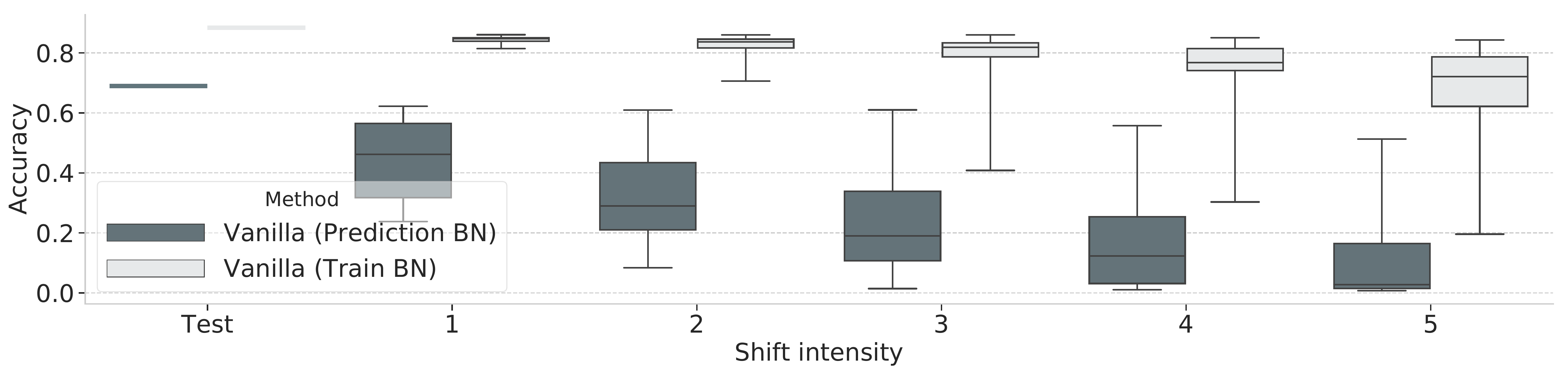}
      \includegraphics[width=0.95\linewidth]{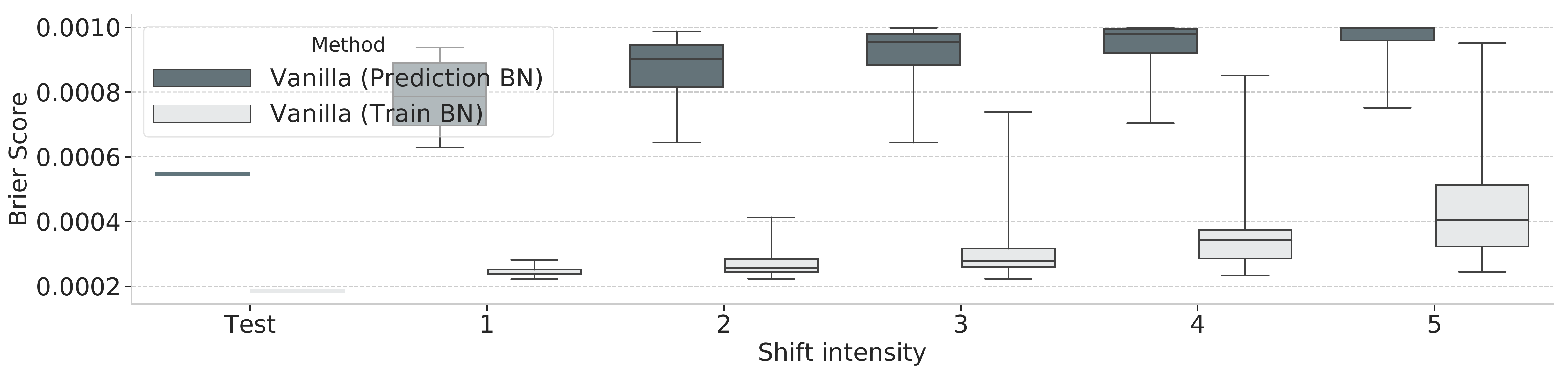}
    \caption{Calibration and accuracy under covariate shift with the ImageNet-C EfficientNet model trained with the Noisy Student technique. Here prediction-time BN actually does worse than train BN, possibly because the massive pre-training of the Noisy Student model exposed the training statistics to many of the shifts encountered in ImageNet-C. Exploring the relationship between pre-training and robustness is an area we will explore in future work.}
    \label{fig:noisy_student}
\end{figure*}

\textbf{Pretraining} We see negative results when evaluating prediction-time BN on a pre-trained Noisy Student model from ~\citep{xie2019self}. As seen in Figure~\ref{fig:noisy_student}, prediction-time BN actually does worse than train BN. We believe this is because the statistics used by train BN contain information from the model pre-training; given that the model was pre-trained on the 300 million images in the JFT~\cite{distillation} dataset, this data likely contains many examples and patterns that resemble the corrupted ImageNet-C data splits. Thus, having this enormous amount of pre-existing information is likely to perform better than using the relatively small amount of shifted data at prediction time. We believe that exploring the relationship between pre-training and model calibration is an exciting area of future work.

\textbf{Natural dataset shifts} While the ImageNet-C benchmark is a popular and challenging dataset, it does not encompass all types of shift typically encountered by a machine learning model in practice. To expand the varieties of covariate shift we evaluate on, we also predict on the ImageNet-v2 test set ~\citep{recht2019imagenet} and ImageNet-A ~\cite{hendrycks2019natural}. ImageNet-v2 is a newly curated test dataset drawn from the same test distribution as ImageNet, and we use the Matched Frequency subet of Imagenet-v2, where the images are sampled to match the same class frequency distributions as the original ImageNet validation dataset. Imagenet-A is a dataset of natural images that have been adversarially curated to minimize classifier accuracy when trained on ImageNet. In Table~\ref{table:imagenet_v2}, we see that while prediction-time BN performs worse on accuracy, it improves calibration as measured by ECE. This accuracy decrease should be expected, because the training EMA statistics used by train BN still accurately represent the activation statistics for this type of change in $p\left(\vx\right)$. However, despite using less accurate normalizing statistics, prediction-time BN still performs competitively. In Table~\ref{table:imagenet_a} prediction-time BN actually outperforms train BN on Imagenet-A, perhaps because the training statistics EMA is not representative of the adversarially constructed test set.
\begin{table}[!ht]
\sisetup{detect-weight,mode=text,group-minimum-digits = 4}
\centering
\bgroup
\def\arraystretch{1.2}
\resizebox{0.7\textwidth}{!}{
\begin{tabular}{|c|c|c|c|}
\hline
\textbf{} &
  {\begin{tabular}[c]{@{}c@{}}Vanilla \\ (Train/Pred BN)\end{tabular}} &
  {\begin{tabular}[c]{@{}c@{}}Ensemble \\ (Train/Pred BN)\end{tabular}} &
  {\begin{tabular}[c]{@{}c@{}}Temp Scaling \\ (Train/Pred BN)\end{tabular}} \\ \hline
Accuracy & \textbf{62.02\%} / 58.32\% & \textbf{65.50\%} / 62.08 \% & \textbf{62.02\%} / 58.32\% \\ \hline
Brier Score &
  \begin{tabular}[c]{@{}c@{}} \textbf{$\num{0.000514}$} /\\ $\num{0.000547}$\end{tabular} &
  \begin{tabular}[c]{@{}c@{}}\textbf{$\num{0.000464}$} /\\ $\num{0.000502}$\end{tabular} &
  \begin{tabular}[c]{@{}c@{}} \textbf{$\num{0.000513}$} /\\ $\num{0.000546}$\end{tabular} \\ \hline
ECE      & 0.085 / \textbf{0.065}     & \textbf{0.020} / 0.026      & 0.080 / \textbf{0.060}     \\ \hline
\end{tabular}
}
\egroup
\vspace{2em}
\caption{Resnet-50 results on ImageNet-v2 for train and prediction-time BN. We see similar behavior of prediction-time BN as on the in-distribution test split of ImageNet, where using the prediction-time statistics instead of the training EMA actually degrades accuracy and Brier Score by a small amount, but at the same time improves ECE. See Figure~\ref{fig:imagenet_methods_comparison_skew_all} for comparisons to the in-distribution test split of ImageNet.}
\label{table:imagenet_v2}
\end{table}

\begin{table}[!ht]
\sisetup{detect-weight,mode=text,group-minimum-digits = 4}
\centering
\resizebox{0.5\textwidth}{!}{
\begin{tabular}{|c|c|c|}
\hline
\textbf{}   & Train BN & Prediction BN \\ \hline
Accuracy    & 1.80\%              & \textbf{1.87\%}                   \\ \hline
Brier Score & $\num{0.001252}$    & \textbf{$\num{0.001148}$}         \\ \hline
ECE         & 0.4070              & \textbf{0.2895}                   \\ \hline
\end{tabular}
}
\vspace{2em}
\caption{DenseNet-121 results on the natural adversarial ImageNet-A dataset for train and prediction-time BN. We see that prediction-time BN outperforms train BN, which could potentially be due to the training EMA statistics not being representative of the adversarial nature of ImageNet-A. We do not use Resnet-50 for these results, as ImageNet-A is adversarially curated to obtain 0\% accuracy with a Resnet-50 architecture. \citep{hendrycks2019natural}}
\label{table:imagenet_a}
\end{table}


\clearpage\newpage

\section{Ablation Studies}\label{sec:ablations}
In addition to measuring the model calibration of prediction-time BN, we also run several studies to explore hypotheses about the cause of its performance improvements. We find that prediction-time BN performs well relative to other normalization techniques and model architectures. Additionally, we closely analyzing model behavior to further investigate the previous hypotheses explaining the improved performance.

\subsection{Sensitivity to $\epsilon$}
An often forgotten parameter in batch normalization is the $\epsilon$ parameter used in the variance term in the denominator (see Equation~\ref{eq:batchnorm}). Originally introduced to avoid division by zero \citep{batchnorm}, it is usually left at its default value\footnote{TensorFlow uses a default of $10^{-3}$ and defines it as a "small float added to variance to avoid dividing by zero"; \url{https://tensorflow.org/versions/r1.15/api_docs/python/tf/keras/layers/BatchNormalization}.} of $10^{-3}$ except for some larger models (such as Resnet-50 on ImageNet) where $\epsilon = 10^{-5}$ is common\footnote{Based on the official TensorFlow model definitions, for example \url{https://git.io/JvsT4}.}.

\begin{figure}[!h]
    \centering
      \includegraphics[width=0.7\linewidth]{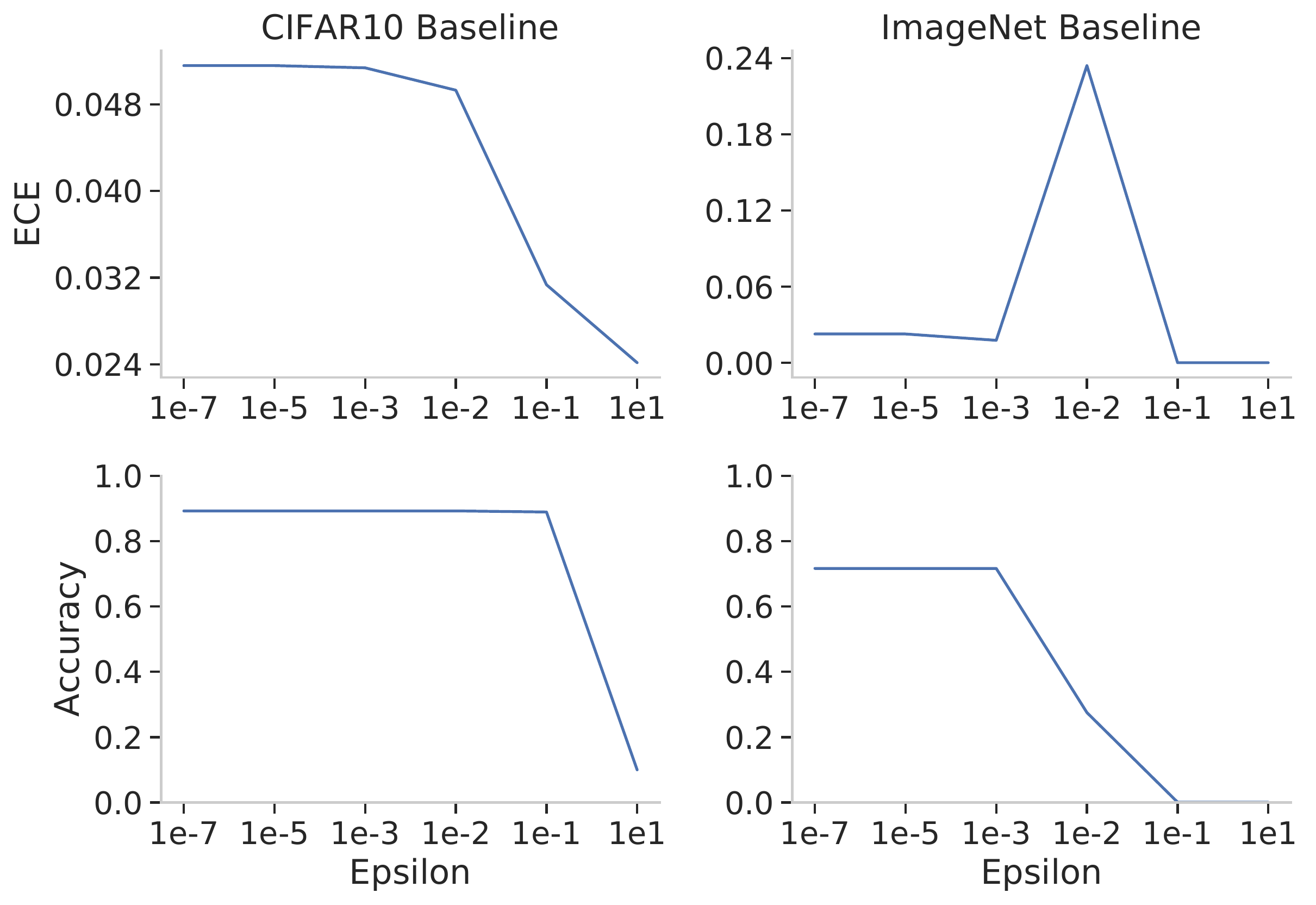}
    \caption{ECE performance across different $\epsilon$ values on the \textbf{in-distribution test set}. The value that performed the best here was chosen to evaluate on the shifted splits. See Figure~\ref{fig:eps_tune_all} for accuracy and Brier score results.}
    \label{fig:eps_tune_compact}
\end{figure}

In the context of model calibration, the denominator term of batch norm can be viewed as temperature scaling that is adaptable through the variance term added to a fixed temperature $\sqrt{\epsilon}$. The classic temperature scaling method  ~\citep{guo2017calibration} selects a temperature that minimizes the negative log-likelihood of the model on the labelled validation set. In contrast, the temperature scaling induced by the batch norm variance is unsupervised and can be adapted per batch. Given that classic temperature scaling with training batch norm does not perform as well as prediction-time BN in Figure~\ref{fig:basic_method_compares}, this adaptive temperature is clearly beneficial. In addition to this, we can follow a similar recipe to ~\citet{guo2017calibration} where we re-tune batch norm's $\epsilon$ for calibration and accuracy on the in-distribution test set and evaluate it on the shifted data. In Figure~\ref{fig:eps_tune_compact} we measure the ECE and accuracy for several values of $\epsilon$ used at prediction time. Given that deep models are typically overly confident under covariate shift ~\citep{ovadia2019}, our intuition is to make $\epsilon$ as large as possible to compensate for this. However, for both models we see $\epsilon$ can only be increased two orders of magnitude from its default values before accuracy collapses. Nonetheless, we can achieve noticeable calibration performance improvements with these higher values. We note that as seen in Figure~\ref{fig:eps_tune_all}, $\epsilon$ does not have much of an effect for ensemble models; one reason for this could be because they are already smoothing the individual models' output distributions when averaging them together. We use the default values of $\epsilon$ for all experiments unless otherwise stated, which is $10^{-3}$ for CIFAR-10 and Criteo and $10^{-5}$ for ImageNet (these are also the values used during training).

\subsection{Comparing Normalization Methods}
\begin{figure}[!h]
    \centering
      \includegraphics[width=0.9\linewidth]{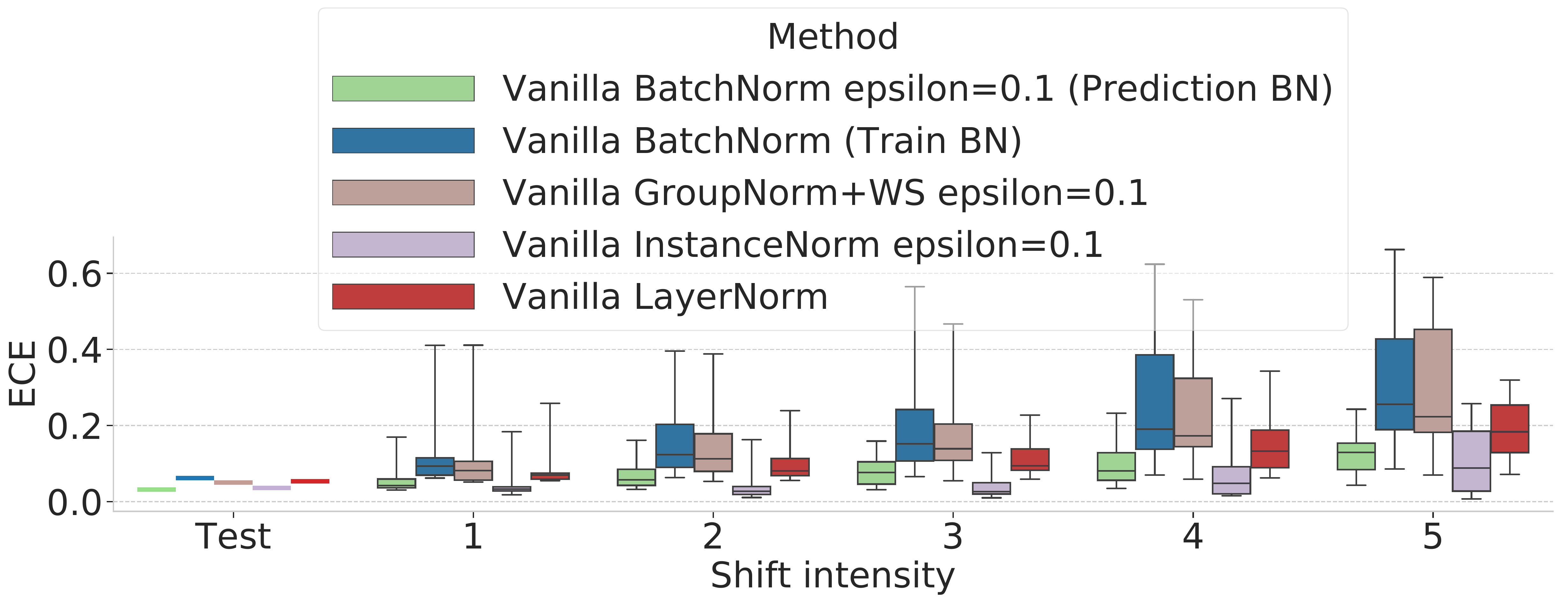}
      \includegraphics[width=0.9\linewidth]{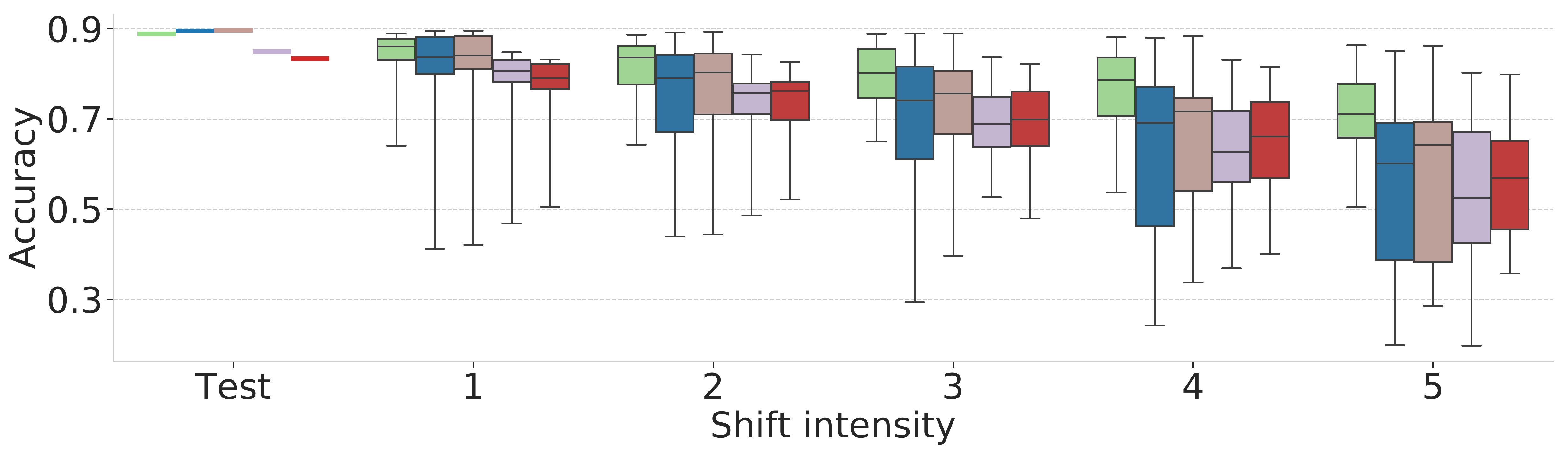}
    \caption{\textbf{CIFAR-10-C vanilla model with different normalization methods.} Each method is the same vanilla model but with the normalization layers changed, and all hyperparameters re-tuned for each. Every method was run with $\epsilon \in \{10^{-3}, 10^{-1}\}$ for batch norm at prediction time, and the best performing runs are included here. While instance norm with $\epsilon=10^{-1}$ has a lower ECE than prediction-time BN, it is notably worse on accuracy. See Figure~\ref{fig:cifar10_bn_methods_all} for more normalization methods.
    }
    \label{fig:cifar10_bn_methods_ece}
\end{figure}

In Figure~\ref{fig:cifar10_bn_methods_ece} we evaluate the calibration and accuracy of several normalization methods on CIFAR-10-C, and the results echo the distribution mismatch seen in Figure~\ref{fig:cifar10_likelihood_ratio_scatter}. Specifically, using the prediction time batch norm significantly improves both calibration and accuracy under shift.  Interestingly, InstanceNorm significantly improves calibration but at the expense of accuracy.

\subsection{Understanding Confidence Distributions}
\begin{figure}[!h]
    \centering
      \includegraphics[width=0.55\linewidth]{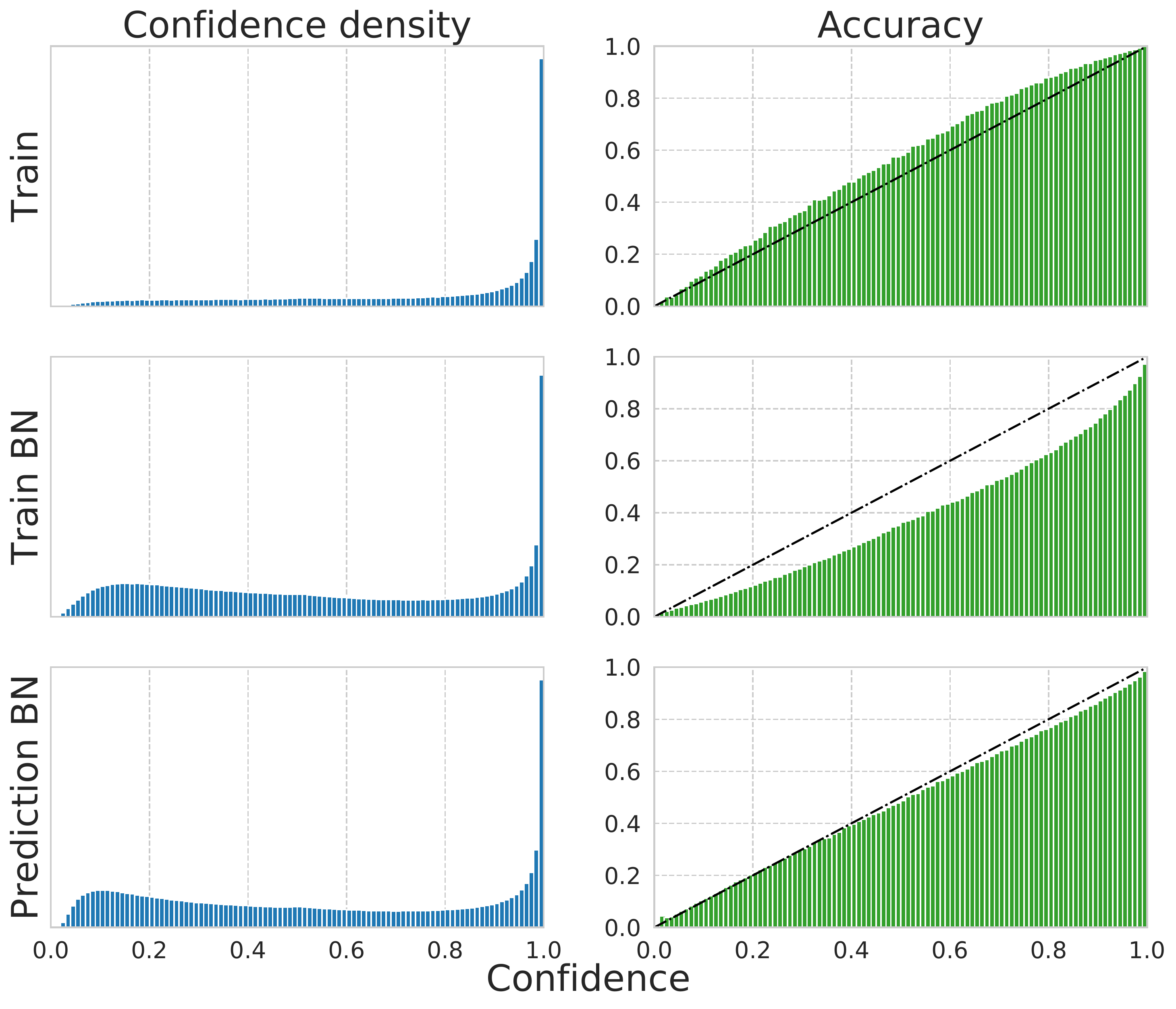}
    \caption{\textbf{Resnet-50 on ImageNet-C confidence distribution and accuracies, grouped into 100 equal width confidence bins.} While the model produces lower confidence predictions on the shifted data regardless of batch norm method, prediction-time BN results in slightly lower confidences and higher per-bin accuracies (but not too high as to be underconfident).}
    \label{fig:imagenet_reliability}
\end{figure}

\begin{figure}[!h]
    \centering
      \includegraphics[width=0.55\linewidth]{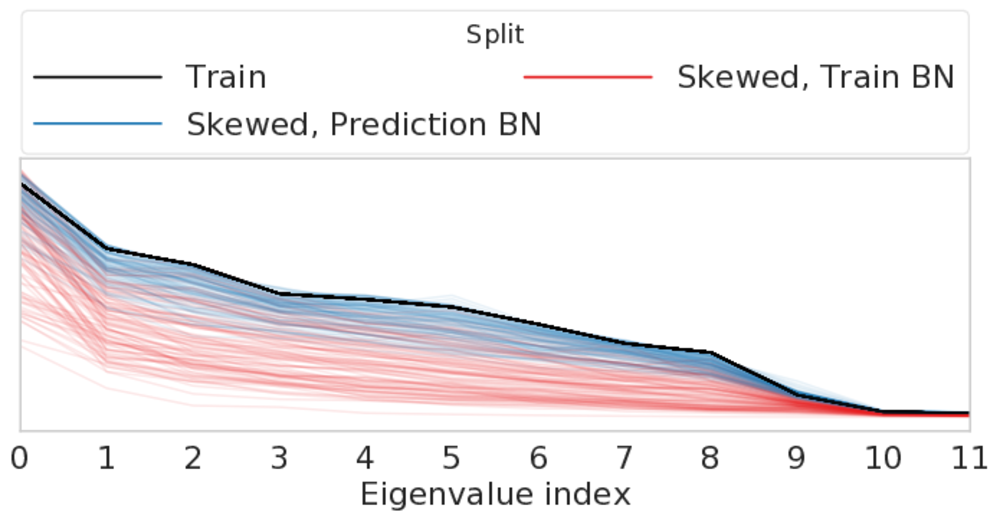}
      \includegraphics[width=0.55\linewidth]{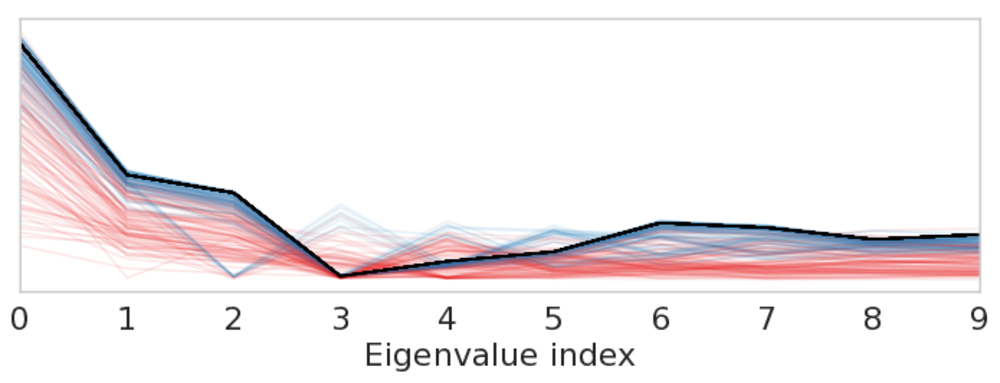}
    \caption{\textbf{The eigenvalues of the covariance matrices for the penultimate layer embeddings (top) and logits (bottom) for Resnet-20 on CIFAR-10.} We analyze the eigenspectrum of these covariance matrices to determine how closely the empirical training and test distributions match.  While these covariance structures do not indicate anything about aligning distribution supports, they are a proxy for how close the distributions are in shape. We expect there to be some change between the training and shifted activation distribution alignments, because the examples are from similar but not identical data distributions. We see prediction-time BN results in a closer but not identical covariance, which could explain the lower confidence compared to training. Additionally, not being too far from the training covariance could also explain the improved accuracy compared to train BN. Note we truncate the eigenspectra after the top 12 eigenvalues for the embeddings because all values after this index were very close to zero.}
    \label{fig:cifar10_cov_eigenvals}
\end{figure}

Given that prediction-time BN is able to map the shifted activations back into the support of the training distribution, we could expect the model to make predictions with similar accuracy and confidence as during training. However, we still see lower confidence predictions with prediction-time BN on ImageNet-C in Figure~\ref{fig:imagenet_reliability}. One possible explanation for this is that even though we are matching the support, we are not precisely matching the distribution densities; while Figure~\ref{fig:cifar10_bn_activation_dists_select} shows that prediction-time BN more closely aligns distribution densities, we see nontrivial misalignment of the eigenspectra of the activation covariance matrices in Figure~\ref{fig:cifar10_cov_eigenvals}.

\clearpage\newpage

\subsection{Batch Normalization Architectures}
While the results in Figure~\ref{fig:cifar10_bn_activation_dists_select} show that prediction-time BN aligns the activation distributions of both the hidden and output layers, we investigate whether or not aligning the layers before the output is necessary to achieve the performance improvements.

\begin{figure}[!h]
    \centering
      \includegraphics[width=0.75\linewidth]{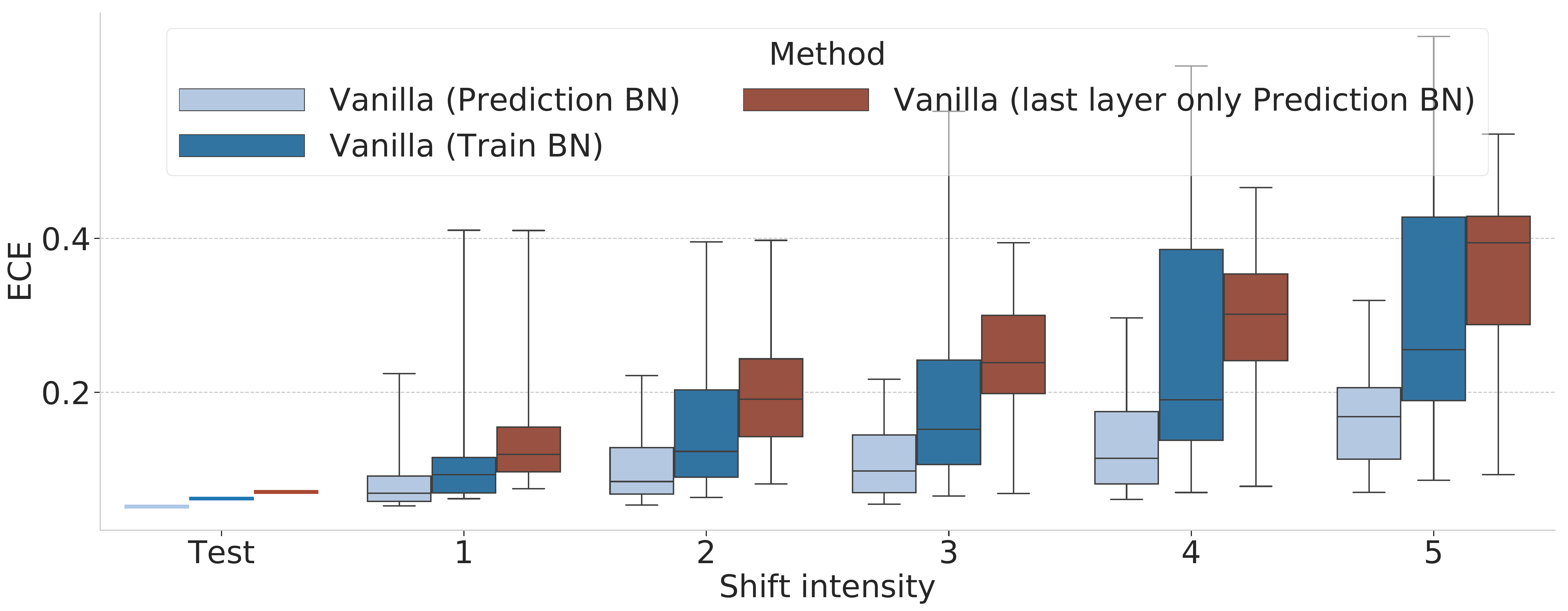}
    \caption{Calibration under covariate shift with the CIFAR-10-C vanilla model where we compare to only using prediction batch statistics on the last batch norm layer. See Figure~\ref{fig:cifar10_test_last_layer_only_all} for more metrics.}
    \label{fig:cifar10_test_last_layer_only_ece}
\end{figure}

\begin{figure}[!h]
    \centering
      \includegraphics[width=0.75\linewidth]{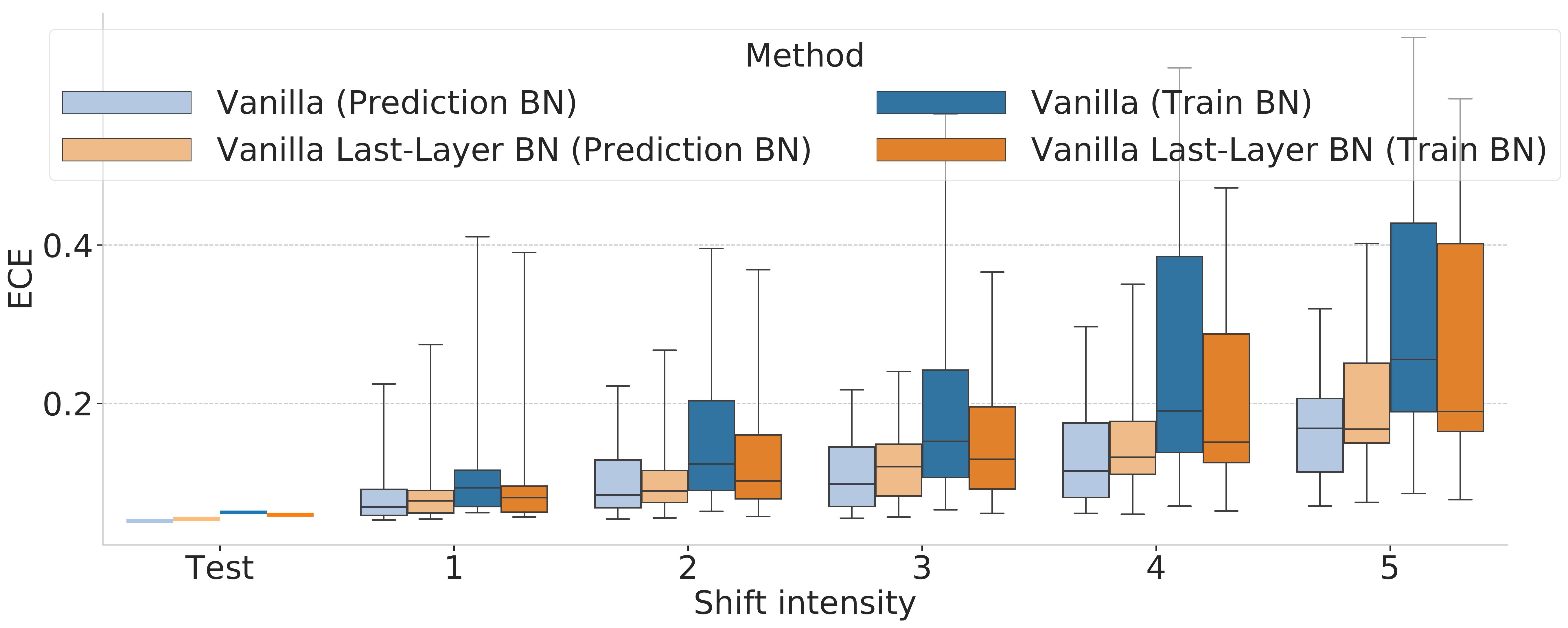}
    \caption{Calibration under covariate shift with the CIFAR-10-C vanilla model compared to an altered Resnet-20 model where we have removed all Batch Norm layers and added one before the final linear layer. We see that we can maintain most, but not all, of the gain in calibration through re-normalizing just the last layer.  See Figure~\ref{fig:cifar10_llbn_compare_all} for more metrics.}
    \label{fig:cifar10_llbn_compare_ece}
\end{figure}

In Figure~\ref{fig:cifar10_test_last_layer_only_ece} we use the training statistics EMA for all batch norm layers except the last, where we use prediction time batch statistics. The model performance noticeably degrades compared to train BN, implying that the final normalization layer alone cannot compensate for the compounded misalignment of all previous hidden layers. We go a step further in Figure~\ref{fig:cifar10_llbn_compare_ece} where we remove all normalization layers except for one immediately before the final linear layer. We can recover most but not all of the performance benefits, implying that using prediction-time BN on the internal normalization layers actively helps improve performance.  However, normalizing the inputs to the last linear layer of the model significantly improves performance under covariate shift.

\vspace{-1em}
\section{Conclusion}
In this paper we propose a cause for deep learning model miscalibration under covariate shift, and offer a simple yet effective remedy. Our hypothesis is that covariate shift causes the internal activations of deep learning models to shift to values outside those encountered during training. A natural solution is to normalize these shifted activations so that they fall within the ranges expected by the model. We explore a variety of techniques to accomplish this, the most effective being what we call prediction-time batch normalization. Our method often outperforms using the training statistic EMA on both image and categorical data modalities, but does not perform as well on more natural dataset shifts and actually decreases performance when combined with pre-training. We leave investigating the relationship between pre-training and activation distribution support mismatch to future work. The requirements for prediction-time batch norm are minimal and realistic for many real world deployment scenarios, as we only need access to a reasonably sized batch of unlabelled data at prediction time.

\pagebreak

\bibliographystyle{plainnat}
\bibliography{references}

\clearpage\newpage
\appendix
\renewcommand{\thefigure}{A\arabic{figure}}
\setcounter{figure}{0}
\setcounter{table}{0}

\section{Datasets}\label{appendix:datasets}

\subsection{CIFAR-10}
For CIFAR-10 training we applied data augmentation as follows: pad by 4 pixels on all sides with zeros, randomly crop to 32x32 pixels, randomly flip the image, and then rescale to be in [-1, 1]. This can be implemented in TensorFlow with the following Python code:

\begin{lstlisting}[language=Python, basicstyle=\small]
image = tf.image.resize_image_with_crop_or_pad(image, 32 + 4, 32 + 4)
image = tf.random_crop(image, [32, 32, 3])
image = tf.image.random_flip_left_right(image)
image = tf.image.convert_image_dtype(image, tf.float32)
image = 2.0 * (image - 0.5)
\end{lstlisting}
For CIFAR-10-C no pre-processing was applied. In addition to the 15 standard corruption types, we also used the extra corruption defined in Appendix B of ~\citep{hendrycks2019benchmarking}, which are \{$\texttt{gaussian\_blur}$, \texttt{saturate}, \texttt{spatter}, \texttt{speckle\_noise}\}. We used the versions of images as provided by TensorFlow Datasets ~\citep{TFDS}.

\subsection{ImageNet}
For ImageNet training we used images of size 224x224, and applied standard Inception data augmentation as defined at this url: \url{https://git.io/JvG6T}.

For ImageNet-C no pre-processing was applied. We used all 19 corruption types, the same as described for CIFAR-10-C.

\subsection{Criteo}
\begin{figure}[!ht]
    \centering
      \includegraphics[width=0.95\linewidth]{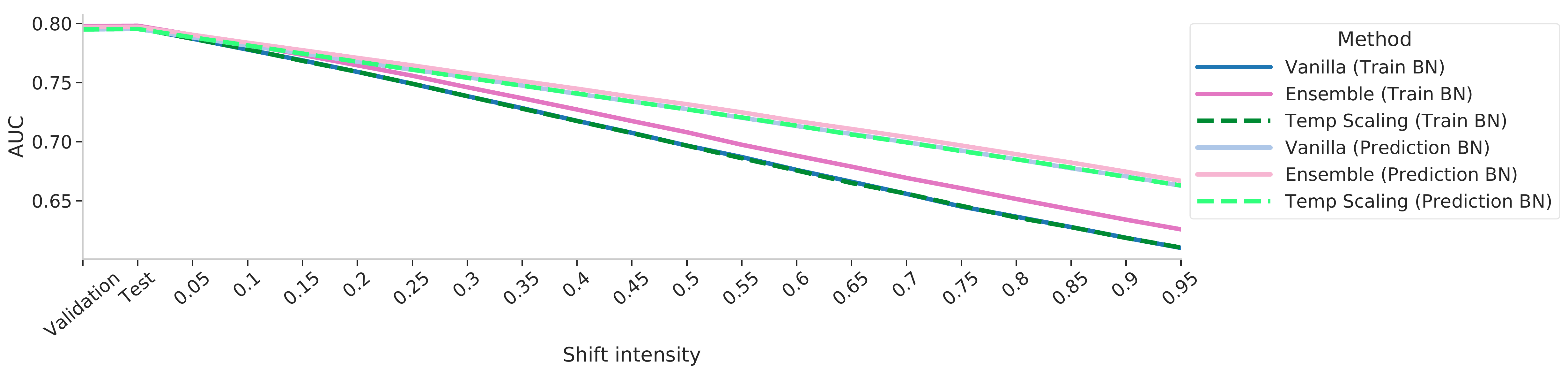}
      \includegraphics[width=0.95\linewidth]{figs/criteo_basic_method_compare_brier_score.pdf}
    \caption{Calibration and AUC under covariate shift on the Criteo dataset.}
    \label{fig:criteo_basic_method_compare_all}
\end{figure}

As done in ~\citep{ovadia2019} we simulate covariate shift in Criteo by randomizing features with increasing probability, ranging from 5\% to 95\% as seen in Figure~\ref{fig:criteo_basic_method_compare_all}.

\section{Models}\label{appendix:models}
\textbf{CIFAR-10} Our CIFAR-10 model is the standard Resnet-20 v1 ~\cite{He2015} with ReLU activations.

\textbf{ImageNet} Our ImageNet model is the standard Resnet-50 v1 ~\cite{He2015} with ReLU activations. For ImageNet-A ~\cite{hendrycks2019natural} we used DenseNet-121 ~\cite{huang2016densely} as defined in tf.keras.applications: \url{https://www.tensorflow.org/api_docs/python/tf/keras/applications/DenseNet121}.

\textbf{Criteo} Our Criteo model is the same as in ~\citep{ovadia2019}. Summarizing, it encodes each categorical feature into a dense vector which are all then concatenated. This feature vector is then fed into a batch normalization layer followed by three fully connected layers of widths $[2572, 1454, 1596]$, each with a ReLU non-linearity.

\section{Hyperparameter Tuning Ranges}\label{appendix:hparams}
Following the recommendations of ~\citet{choi2019empirical}, we use random search within the ranges defined in Table~\ref{table:hparam_ranges} to tune all available hyperparameters for each optimizer. We used 100 random trials for all experiments to tune the learning rate $\alpha$, one minus the momentum $1 - \gamma$, and Adam's $\epsilon$ on a logarithmic scale.

\begin{table}[!h]
\centering
\newcommand{\srange}[2] {\lbrack\num{#1}, \num{#2}\rbrack}
\begin{tabular}{|c|c|c|c|c|}
\hline
Experiment & Optimizer & $\alpha$ & $1 - \gamma$ & $\epsilon$ \\ \hline
CIFAR-10 Resnet-20 & Adam & $\srange{e-3}{e0}$ & $\srange{e-2}{0.15}$ & $\srange{e-8}{e-5}$ \\ \hline
ImageNet Resnet-50 & Nesterov & $\srange{5e-3}{5e-1}$ & $\srange{e-3}{0.15}$ & --- \\ \hline
Criteo MLP & Adam & $\srange{e-4}{e-1}$ & $\srange{e-2}{0.15}$ & $\srange{e-8}{e-5}$ \\ \hline
\end{tabular}
\label{table:hparam_ranges}
\caption{Tuning ranges for each hyperparameter.}
\end{table}

For CIFAR-10, we trained for 100 epochs with a batch size of 512. We used a learning rate schedule where the learning rate was reduced at epochs ${40, 60, 80, 90}$ by ${0.1, 0.01, 0.001, 0.0005}$.

For ImageNet, we trained for 90 epochs with a batch size of 512. We used a learning rate schedule where the learning rate started at 0 and was linearly increased for the first 5 epochs to $\alpha$, then reduced at epochs ${30, 60, 80}$ by ${0.1, 0.01, 0.001}$.

For Criteo, we trained for 1 epoch with a batch size of 1024 and the same learning rate schedule used for CIFAR-10.

\section{Additional Figures}\label{appendix:figures}

\begin{figure}[!ht]
    \centering
  \includegraphics[width=0.95\linewidth]{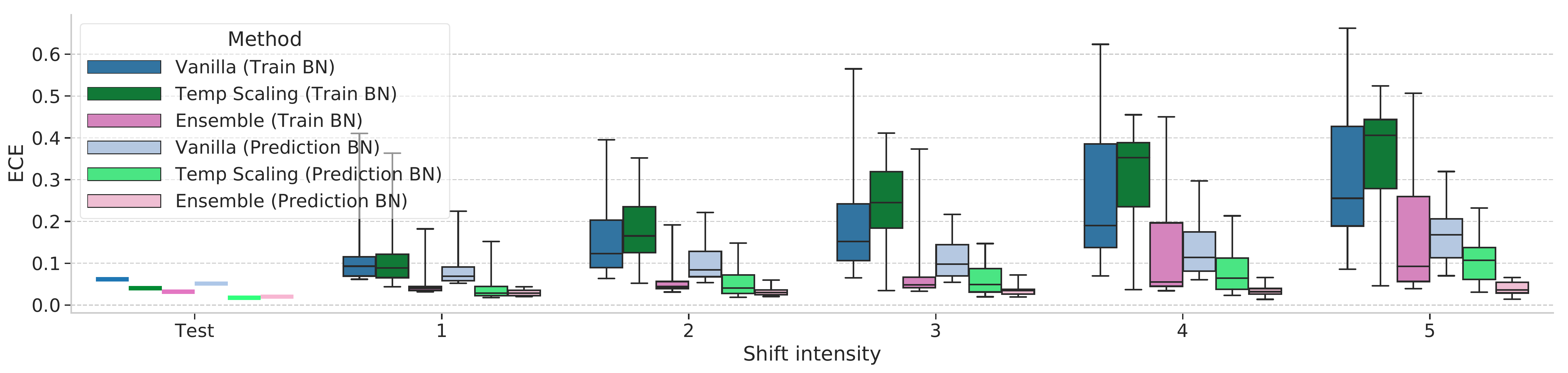}
      \includegraphics[width=0.95\linewidth]{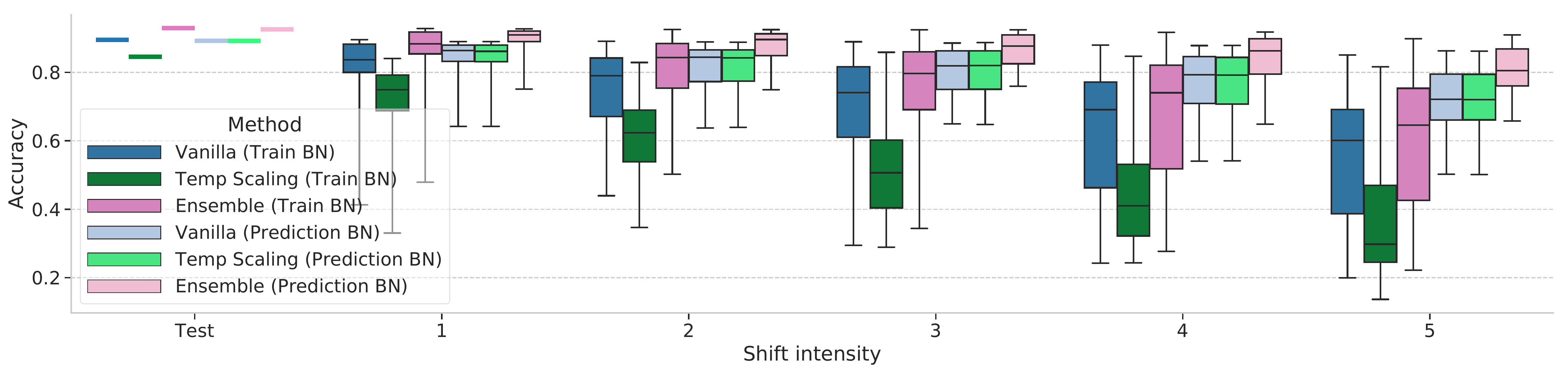}
      \includegraphics[width=0.95\linewidth]{figs/cifar10_basic_method_compare_brier_score}
    \caption{Calibration and accuracy under covariate shift on CIFAR-10-C for vanilla, ensemble, and temperature scaling methods, each with a test batch size of 500 and $\epsilon = 10^{-3}$.}
    \label{fig:cifar10_methods_comparison_skew_all}
\end{figure}

\begin{figure}[!ht]
    \centering
      \includegraphics[width=0.85\linewidth]{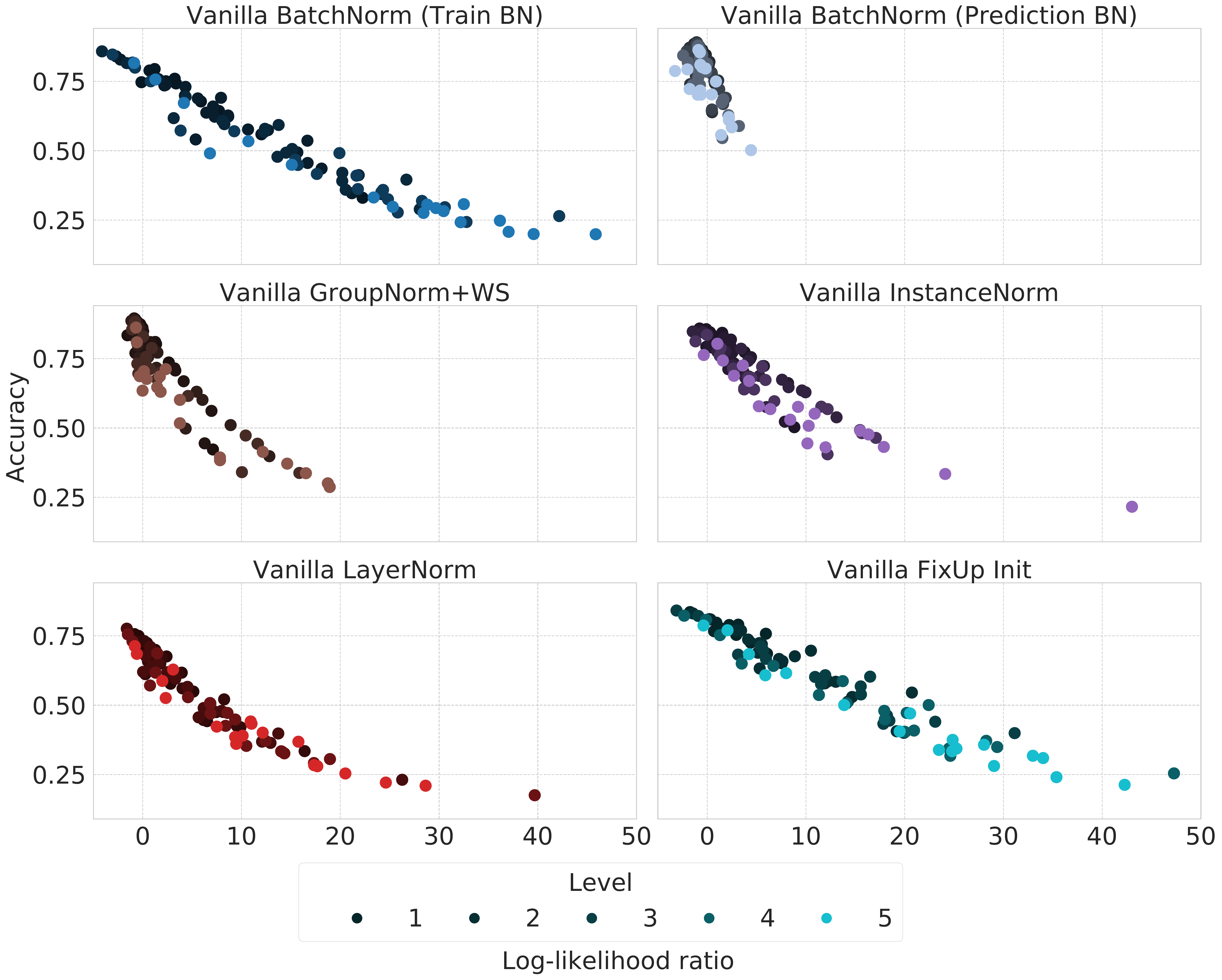}
      \includegraphics[width=0.85\linewidth]{figs/cifar10_likelihood_ratio_scatter_norm_types_by_split_brier_score.pdf}
    \caption{As first described in Figure~\ref{fig:cifar10_likelihood_ratio_scatter}, we also see a linear trend of degrading performance as the train and test activation distributions become further apart. Once again, higher shift intensities are plotted with higher color saturation, illustrating a clear relationship between shift intensity and distance between activation supports.
    }
    \label{fig:cifar10_likelihood_ratio_scatter_all}
\end{figure}

\begin{figure}[!ht]
    \centering
      \includegraphics[width=0.95\linewidth]{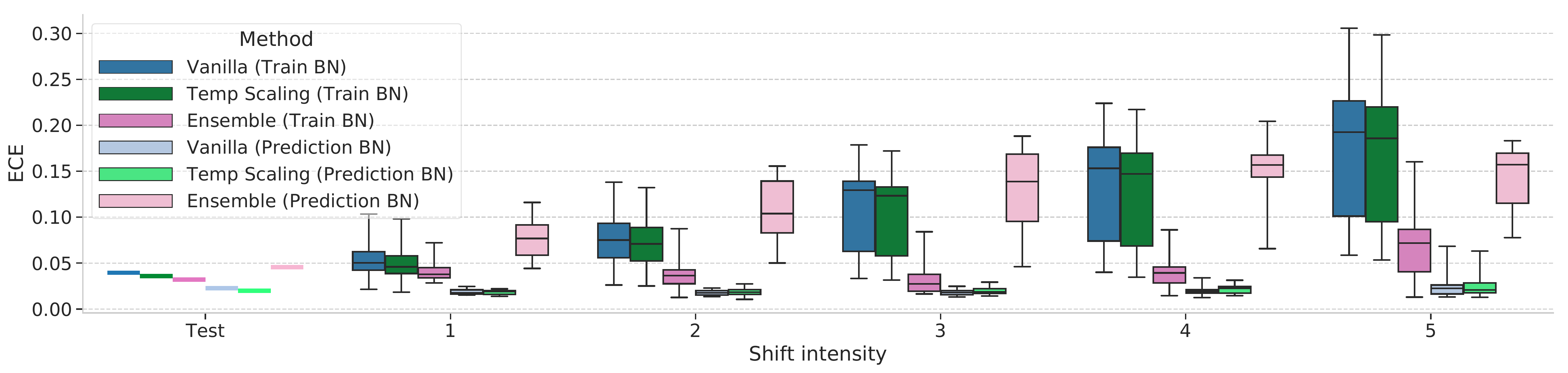}
      \includegraphics[width=0.95\linewidth]{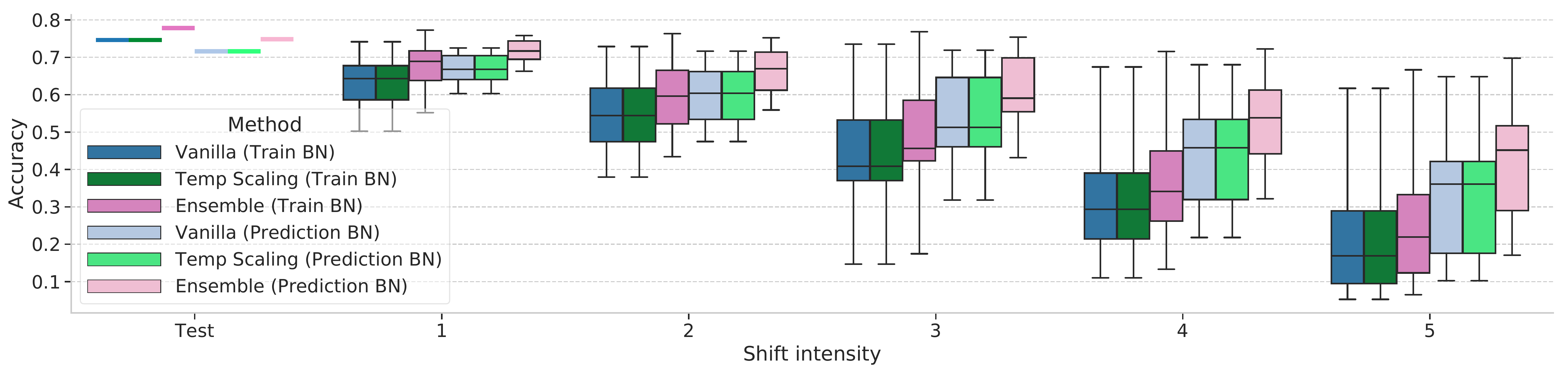}
      \includegraphics[width=0.95\linewidth]{figs/imagenet_basic_method_compare_brier_score}
    \caption{Calibration and accuracy under covariate shift on ImageNet-C for vanilla, ensemble, and temperature scaling methods, each with a test batch size of 100 and $\epsilon = 10^{-5}$. Note that for ensembles using prediction-time BN actually degrades performance on ECE, but  achieves the best accuracy and Brier Score. This is likely due to an issue with the number of bins used to compute ECE, which was 30 for all ImageNet experiments, because we still see improvements on Brier Score which, unlike ECE, is a proper scoring rule.}
    \label{fig:imagenet_methods_comparison_skew_all}
\end{figure}

\begin{figure}[!ht]
    \centering
      \includegraphics[width=0.95\linewidth]{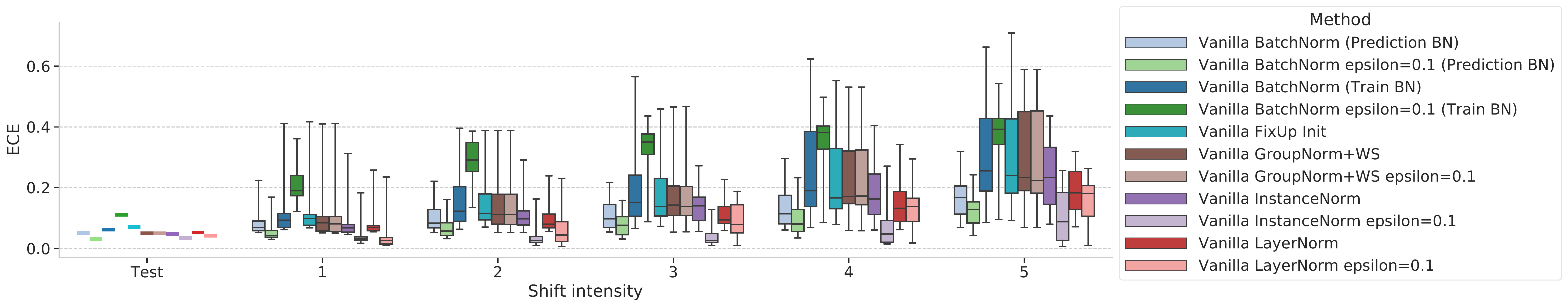}
      \includegraphics[width=0.95\linewidth]{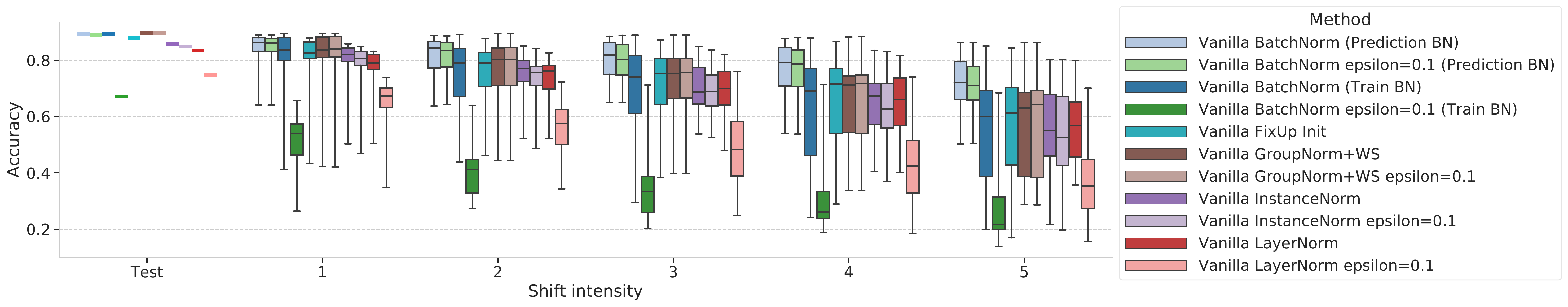}
      \includegraphics[width=0.95\linewidth]{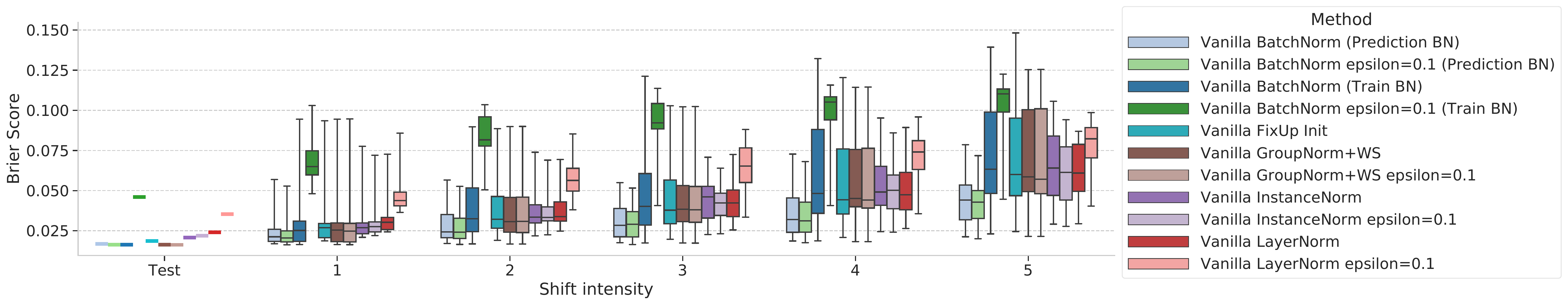}
    \caption{Calibration and accuracy under covariate shift on CIFAR-10-C for the vanilla model using various normalization techniques, each with a test batch size of 500. In addition to $\epsilon \in \{10^{-3}, 10^{-1}\}$ varieties of each normalization method, we also include FixUp initialization ~\citep{zhang2019fixup} as a no normalization baseline. FixUp performs well on the in-distribution set but quickly degrades in a similar trend as the train BN method.}
    \label{fig:cifar10_bn_methods_all}
\end{figure}

\begin{figure}[!ht]
    \centering
      \includegraphics[width=0.95\linewidth]{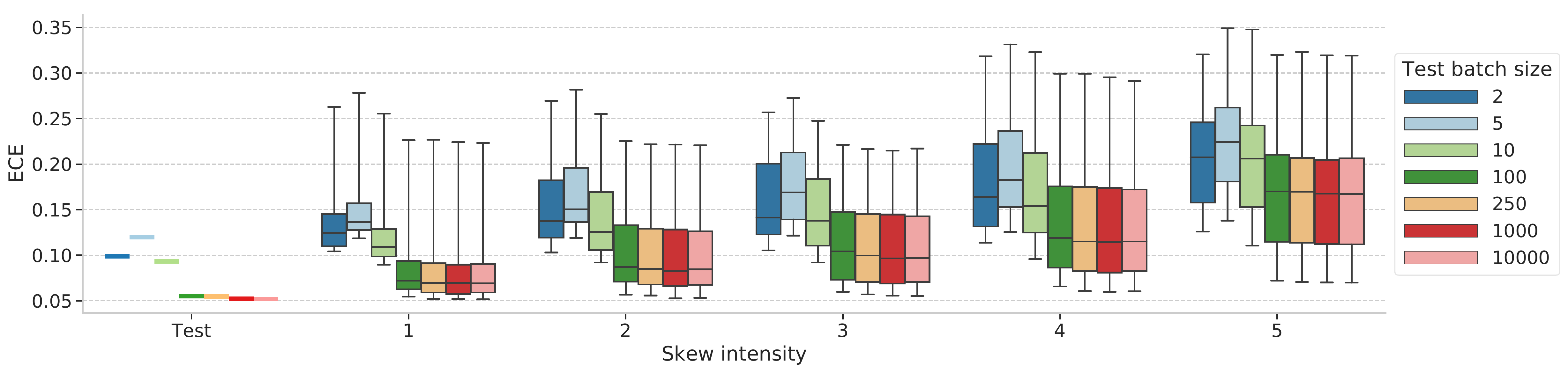}
      \includegraphics[width=0.95\linewidth]{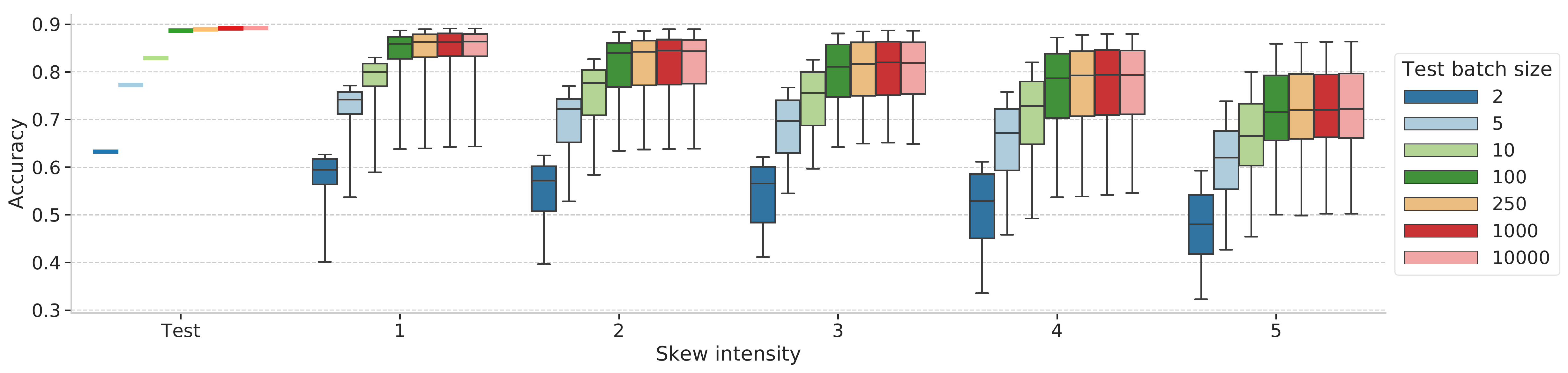}
      \includegraphics[width=0.95\linewidth]{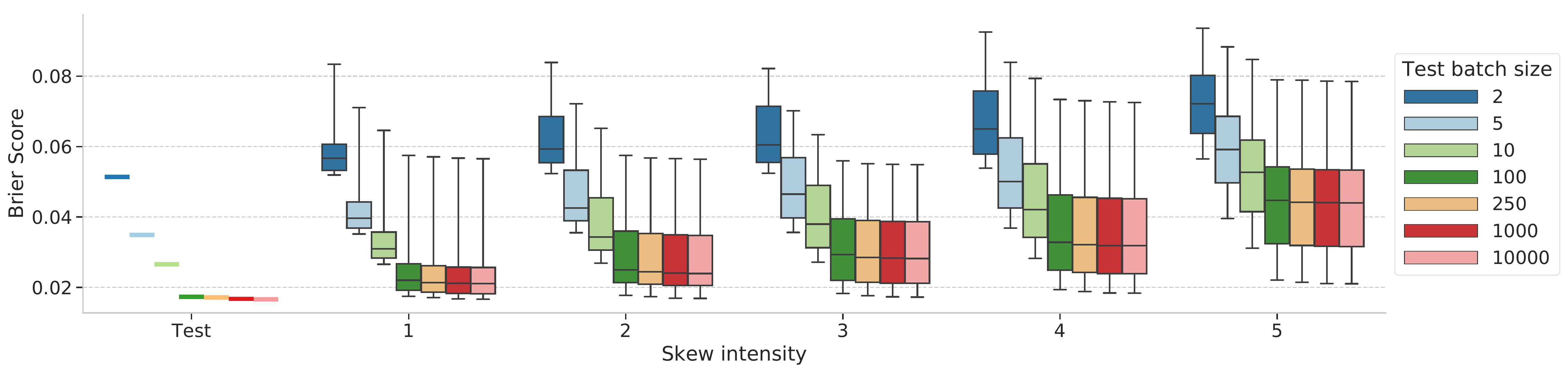}
    \caption{CIFAR-10-C calibration and accuracy for the vanilla model for different test batch sizes.}
    \label{fig:cifar_test_bs_all}
\end{figure}

\begin{figure}[!ht]
    \centering
      \includegraphics[width=0.95\linewidth]{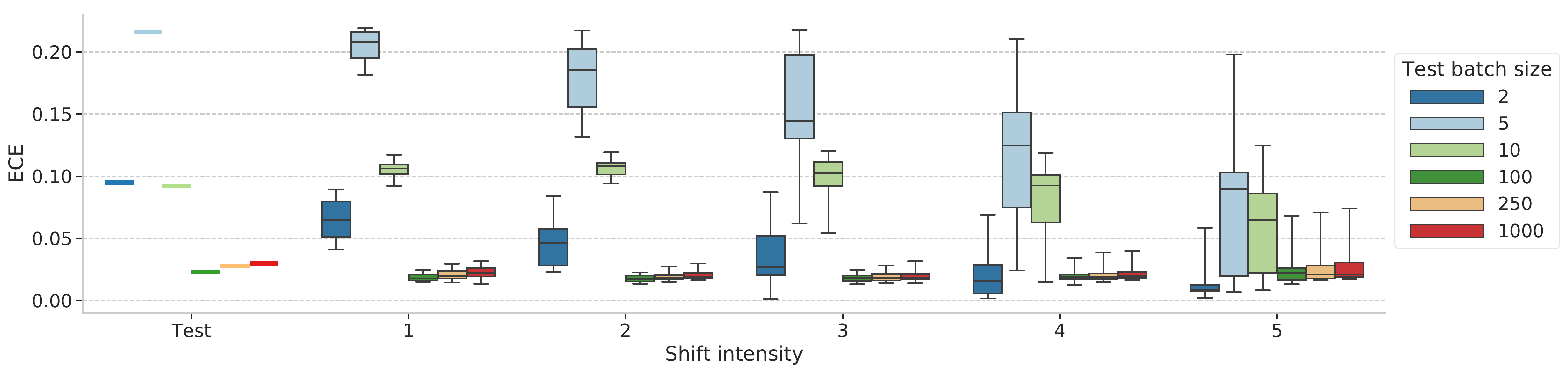}
      \includegraphics[width=0.95\linewidth]{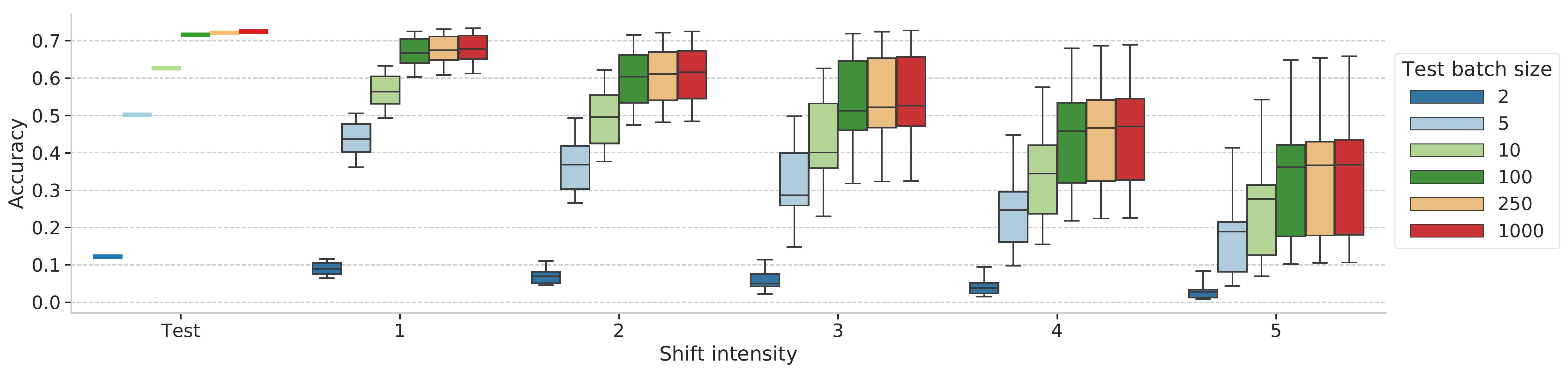}
      \includegraphics[width=0.95\linewidth]{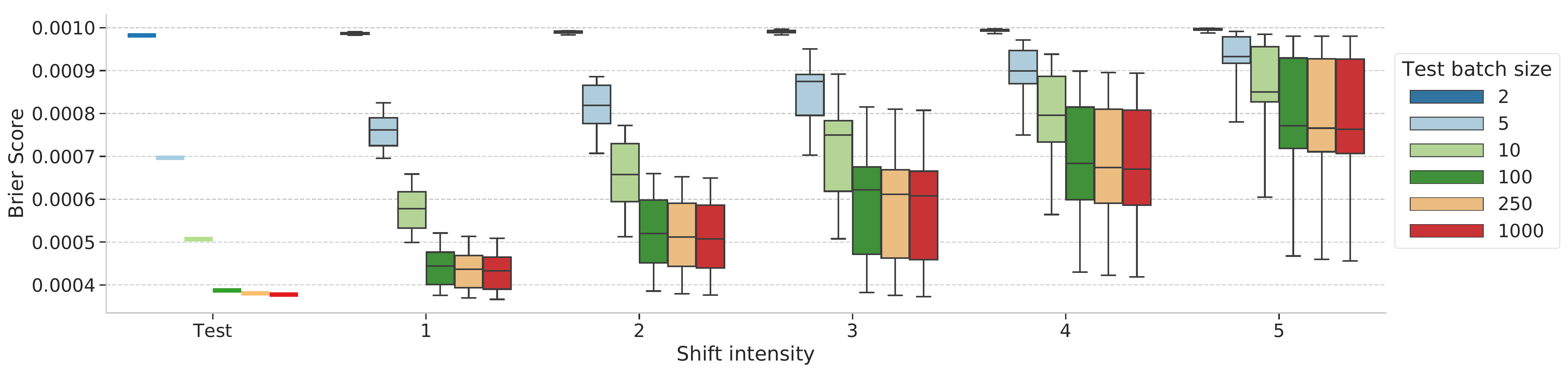}
    \caption{ImageNet-C calibration and accuracy for the vanilla model for different test batch sizes. We see an almost identical trend as in CIFAR-10-C, with performance plateauing after batch size 100 or 250. We do see a small increase in ECE after batch size 100, but accuracy and Brier Score continue to slightly improve, so this could be an artifact of how ECE bins confidences.}
    \label{fig:imagenet_test_bs_all}
\end{figure}

\begin{figure}[!ht]
    \centering
      \includegraphics[width=0.95\linewidth]{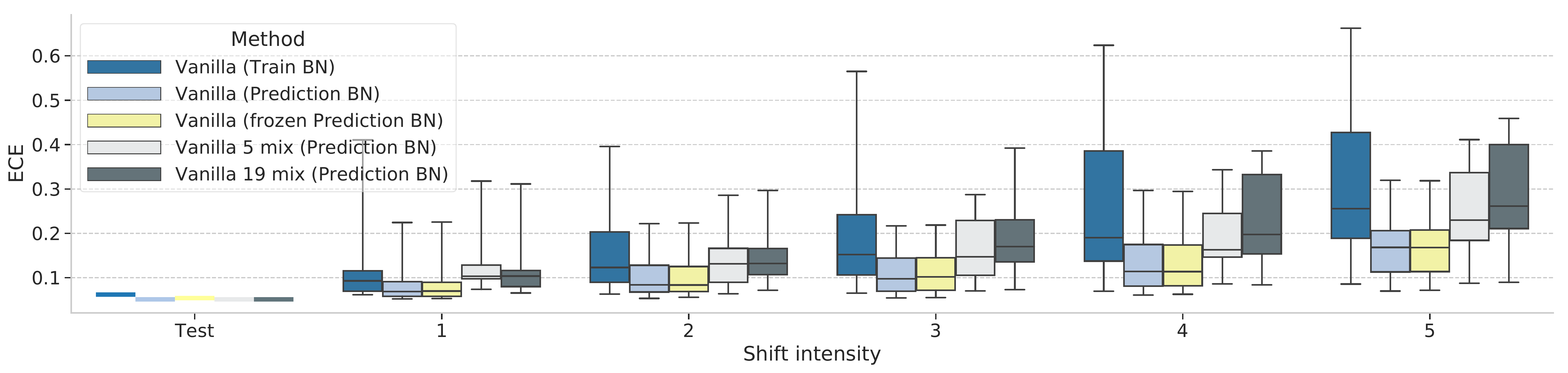}
      \includegraphics[width=0.95\linewidth]{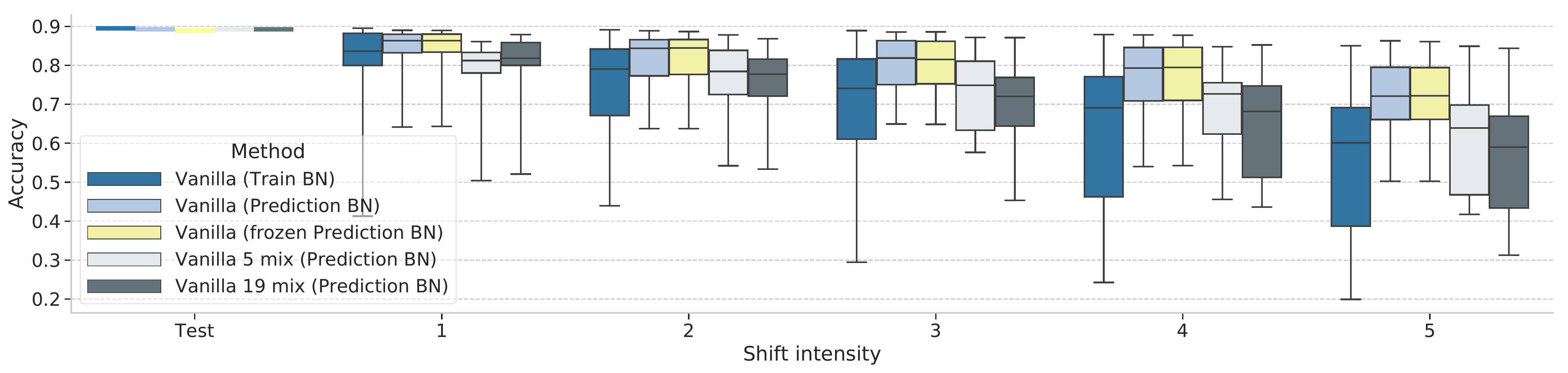}
      \includegraphics[width=0.95\linewidth]{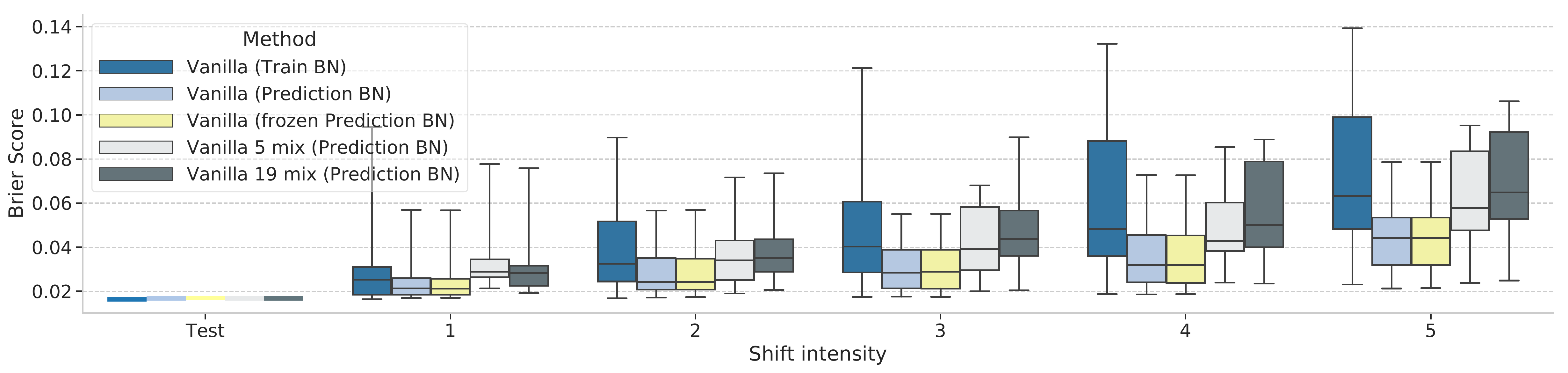}
    \caption{Calibration and accuracy on CIFAR-10-C. For frozen prediction-time BN we use the test batch statistics from the first batch on the whole data split, and for 5 and 19 mix prediction-time BN we mix together 5 and 19 different shifts simultaneously to test the robustness of prediction-time BN.}
    \label{fig:cifar_bn_stat_method_all}
\end{figure}

\begin{figure}[!ht]
    \centering
      \includegraphics[width=0.95\linewidth]{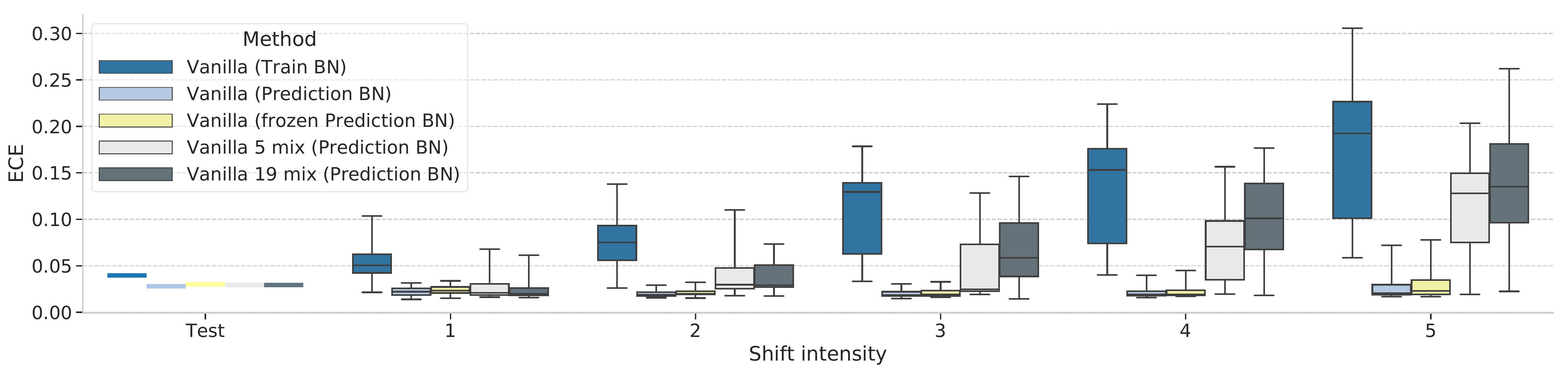}
      \includegraphics[width=0.95\linewidth]{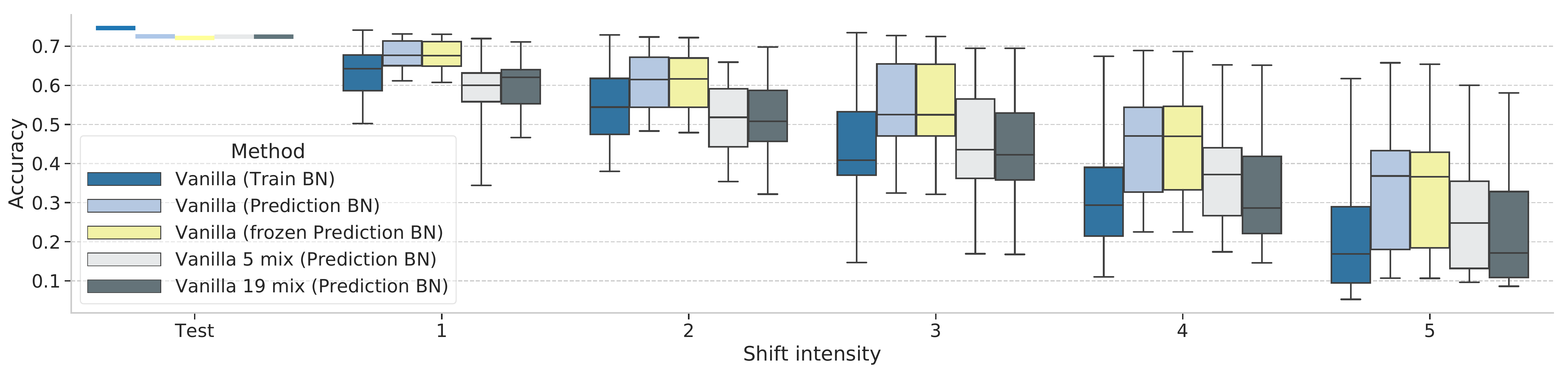}
      \includegraphics[width=0.95\linewidth]{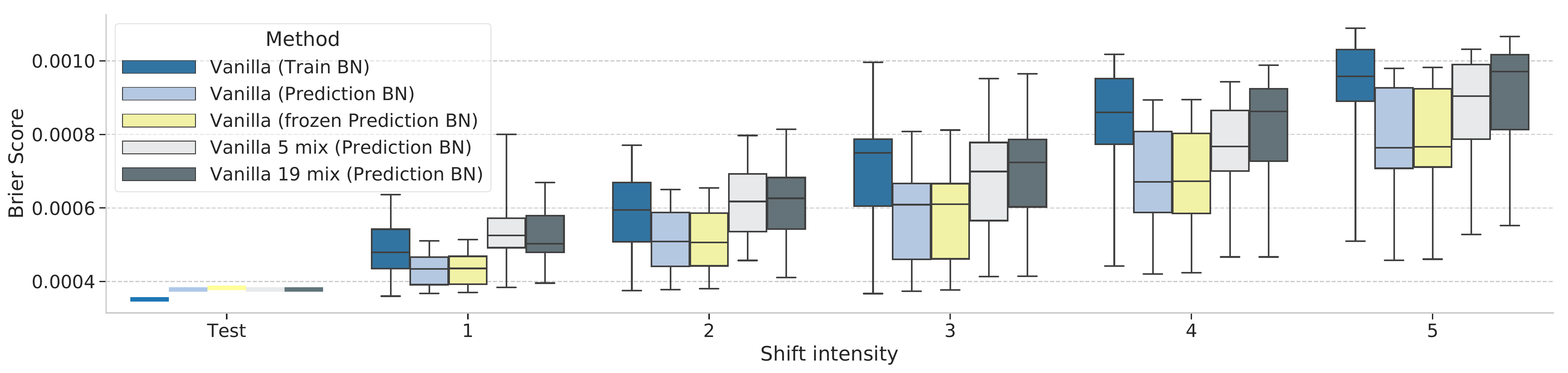}
    \caption{The same calibration and accuracy experiments as Figure~\ref{fig:cifar_bn_stat_method_all} but on ImageNet-C, with a test batch size of 500.}
    \label{fig:imagenet_bn_stat_method_all}
\end{figure}


\begin{figure}[!ht]
    \centering
      \includegraphics[width=0.95\linewidth]{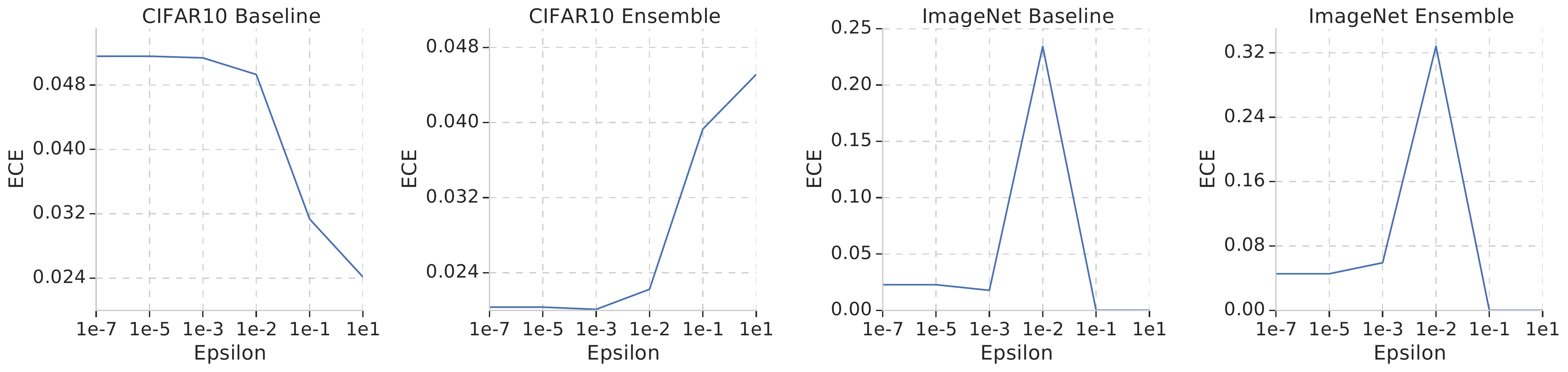}
      \includegraphics[width=0.95\linewidth]{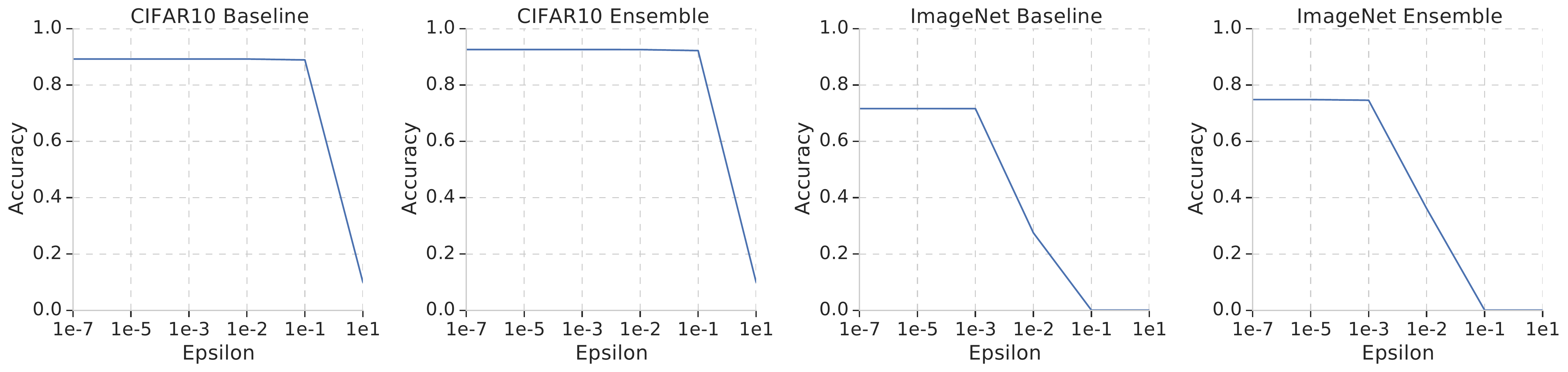}
      \includegraphics[width=0.95\linewidth]{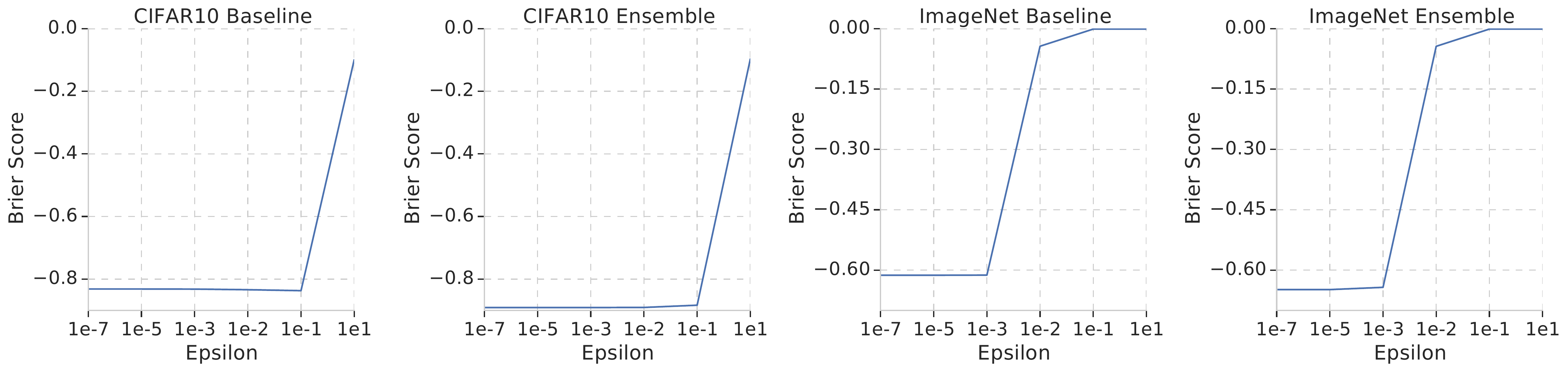}
    \caption{Calibration and accuracy across different $\epsilon$ values for CIFAR-10 and ImageNet on the \textbf{in-distribution test set}. The best performing values were chosen to evaluate on the skewed sets. Note that for ensembles on both CIFAR-10 and ImageNet, increasing $\epsilon$ from their default values only hurts performance, while for single models we can get noticeable ECE performance improvements while maintaining accuracy and Brier Score.}
    \label{fig:eps_tune_all}
\end{figure}

\begin{figure}[!ht]
    \centering
      \includegraphics[width=0.95\linewidth]{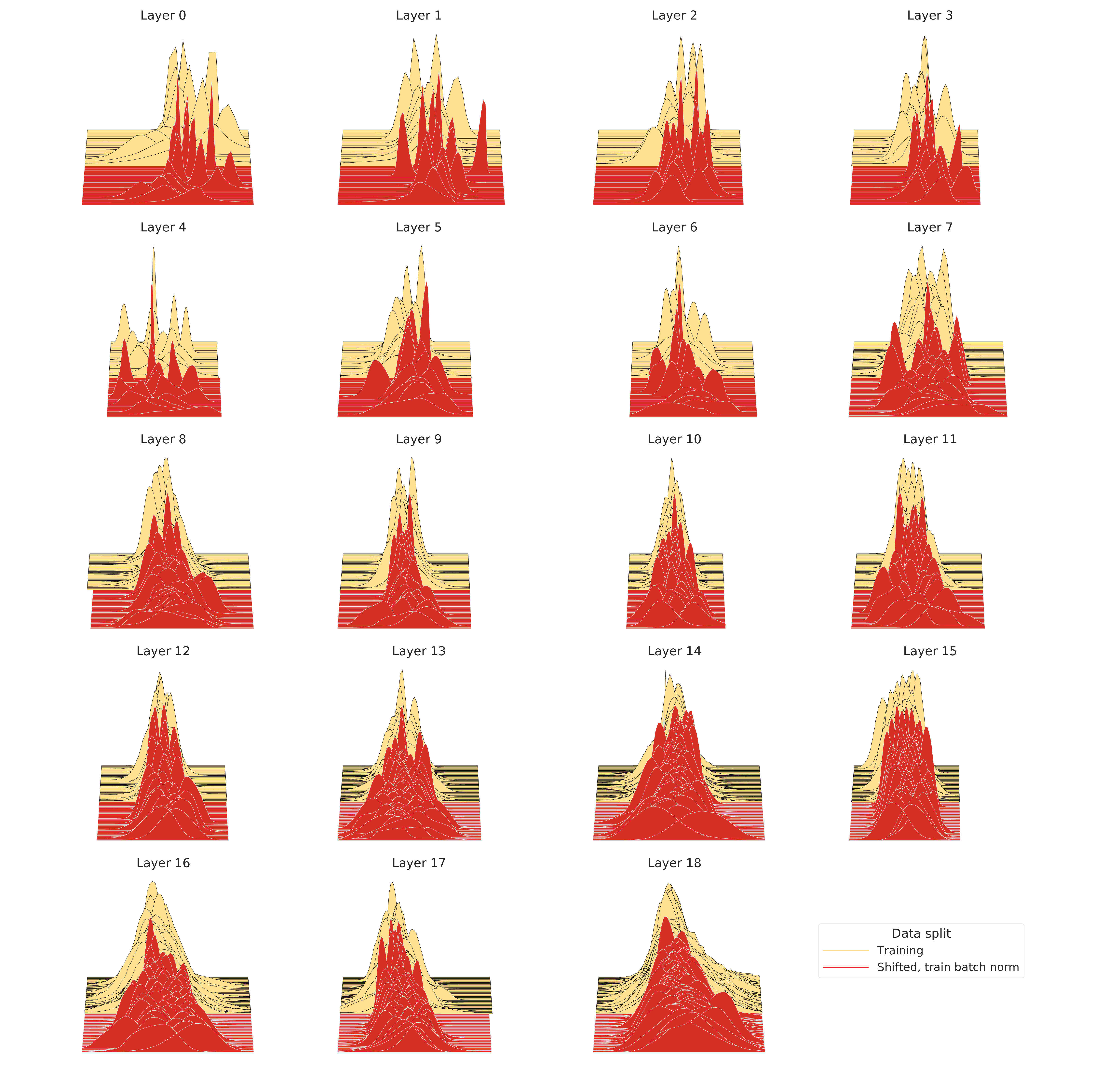}
    \caption{Empirical distributions of the outputs of each normalization layer in Resnet-20, for train BN.}
    \label{fig:cifar10_bn_activation_dists_ema_all}
\end{figure}

\begin{figure}[!ht]
    \centering
      \includegraphics[width=0.95\linewidth]{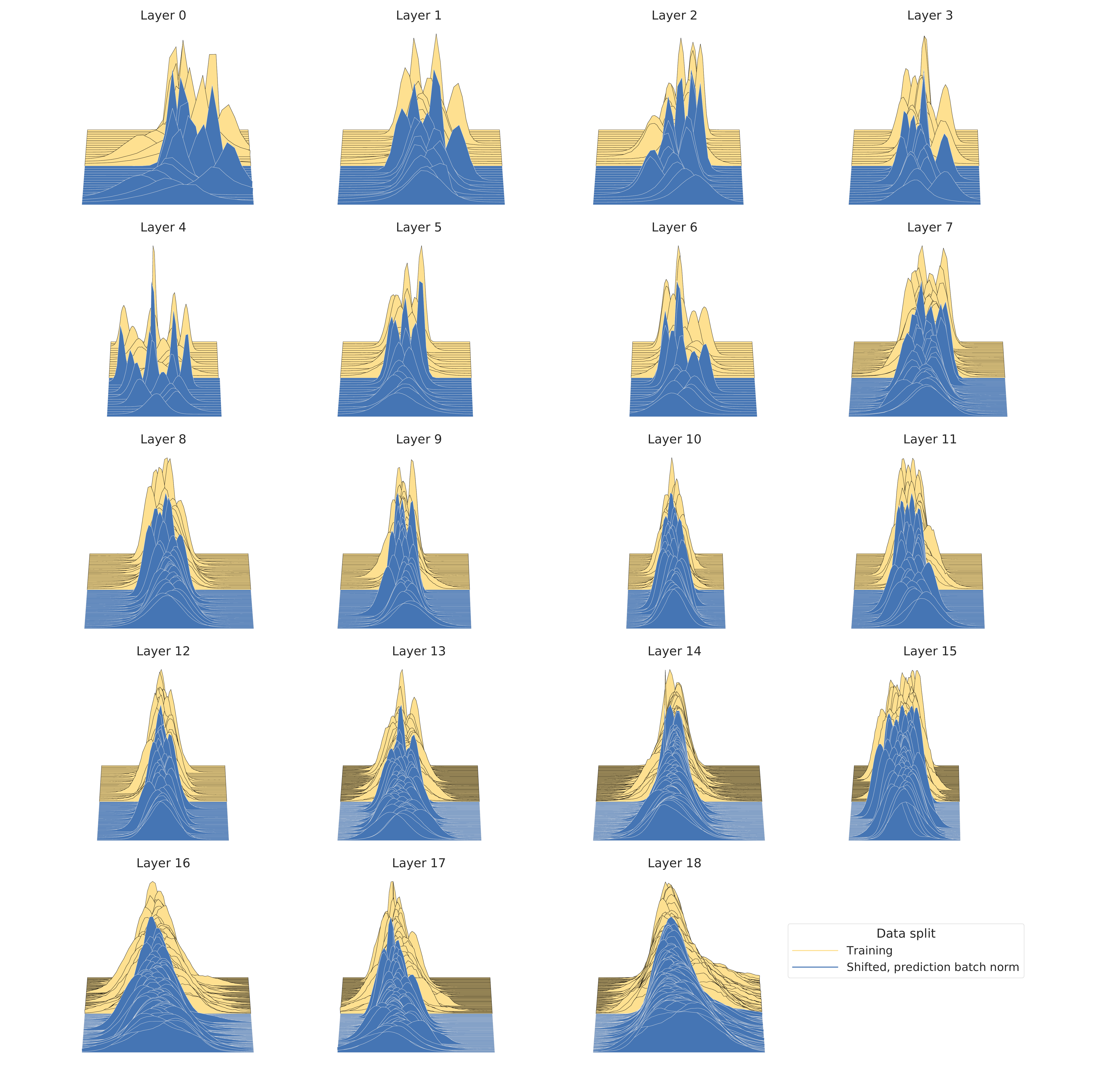}
    \caption{Empirical distributions of the outputs of each normalization layer in Resnet-20, for prediction-time BN.}
    \label{fig:cifar10_bn_activation_dists_test_batch_all}
\end{figure}

\begin{figure}[ht]
    \centering
      \includegraphics[width=0.95\linewidth]{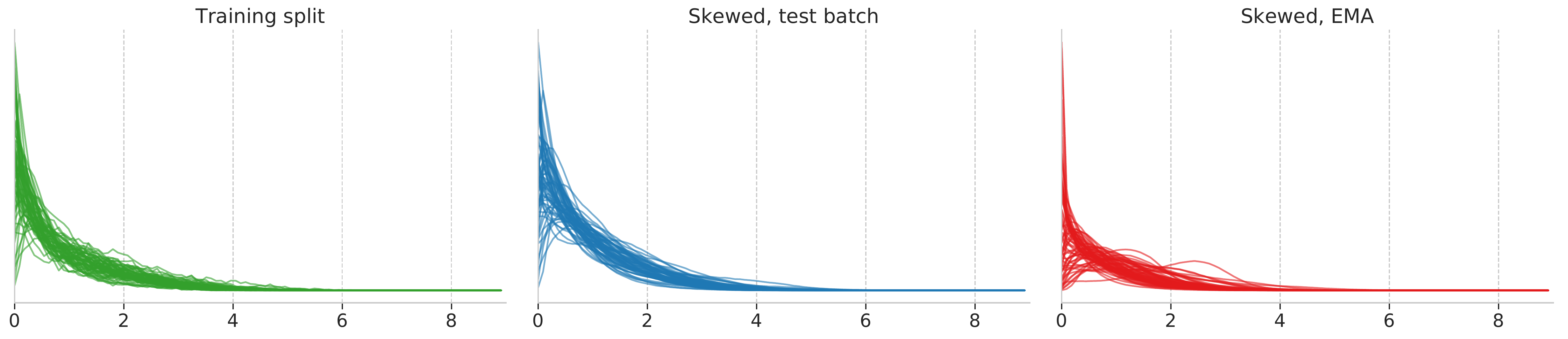}
      \includegraphics[width=0.95\linewidth]{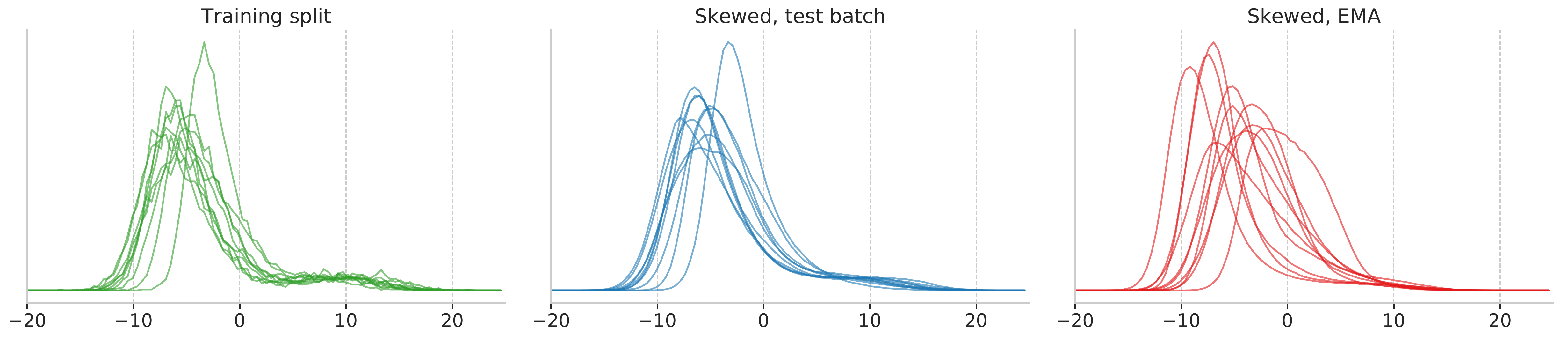}
    \caption{Per-dimension distributions for the penultimate layer embeddings and logits of Resnet-20 on CIFAR-10. Similar to Figure~\ref{fig:cifar10_bn_activation_dists_select}, we see prediction-time BN more effectively aligns both activation supports and marginal distributions. Additionally, train BN has many logit values that are larger in magnitude, which supports the trend of more confident predictions like those in Figure~\ref{fig:imagenet_reliability}.}
    \label{fig:cifar10_bn_final_dists}
\end{figure}

\begin{figure}[!ht]
    \centering
      \includegraphics[width=0.95\linewidth]{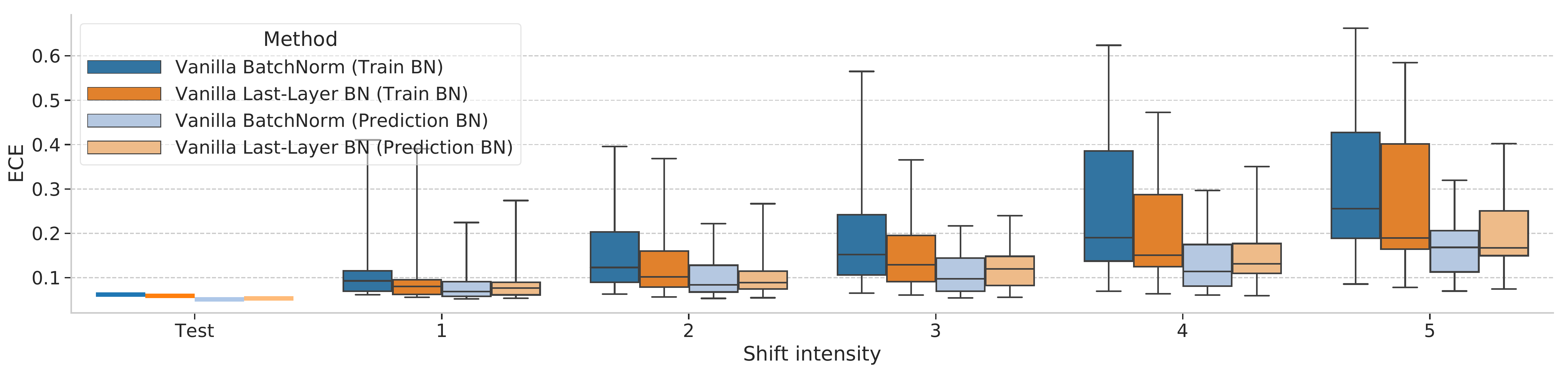}
      \includegraphics[width=0.95\linewidth]{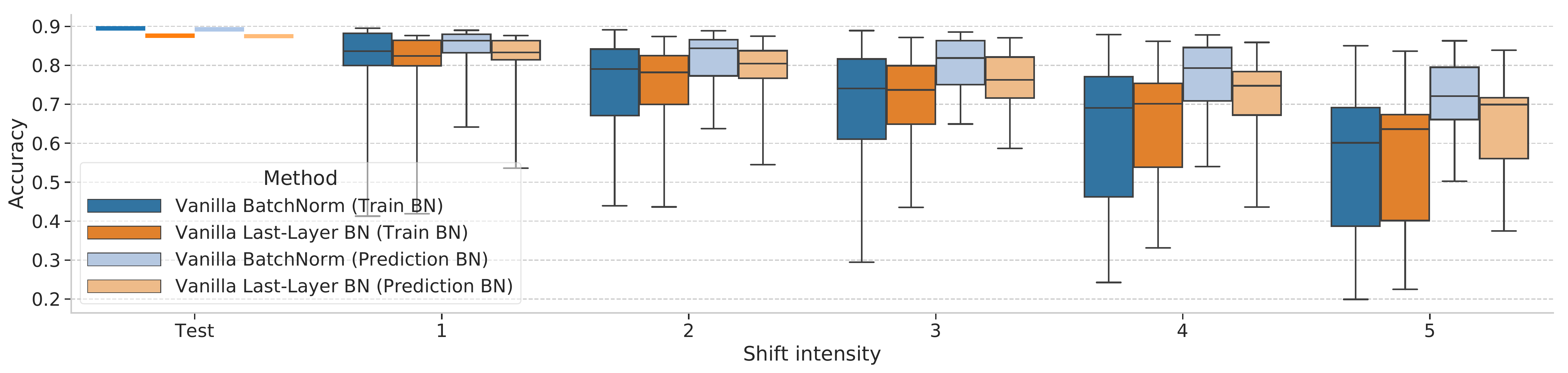}
      \includegraphics[width=0.95\linewidth]{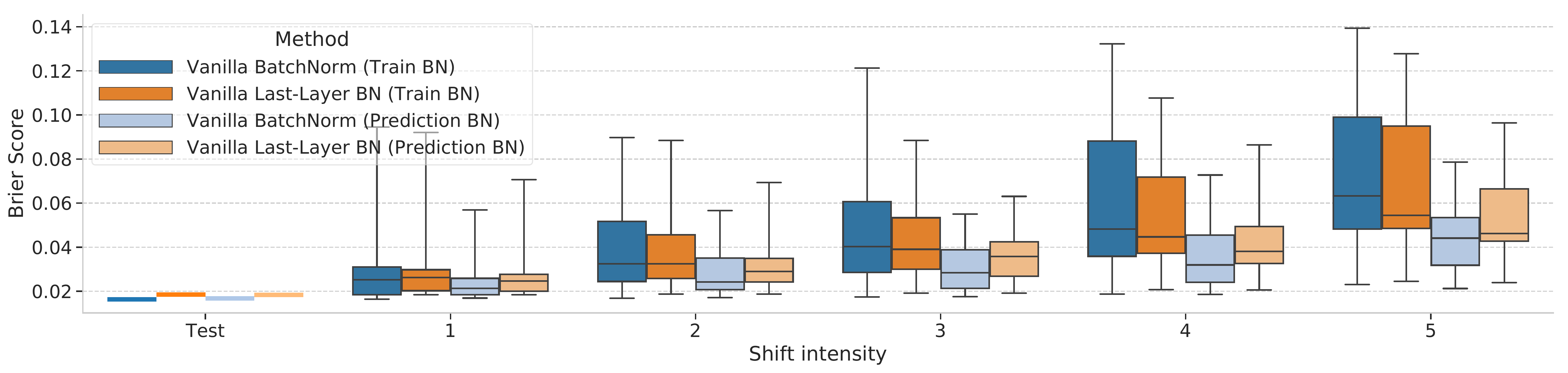}
    \caption{Calibration and accuracy under covariate shift on CIFAR-10-C, for the vanilla model and  an altered Resnet-20 model where all batch norm layers were removed and one was added before the final linear layer.}
    \label{fig:cifar10_llbn_compare_all}
\end{figure}

\begin{figure}[ht]
    \centering
      \includegraphics[width=0.95\linewidth]{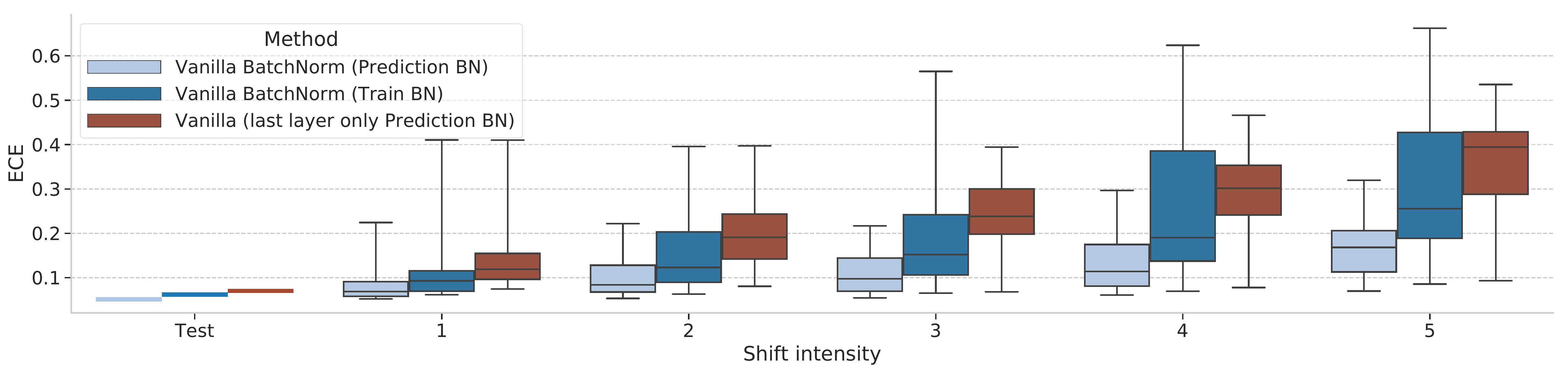}
      \includegraphics[width=0.95\linewidth]{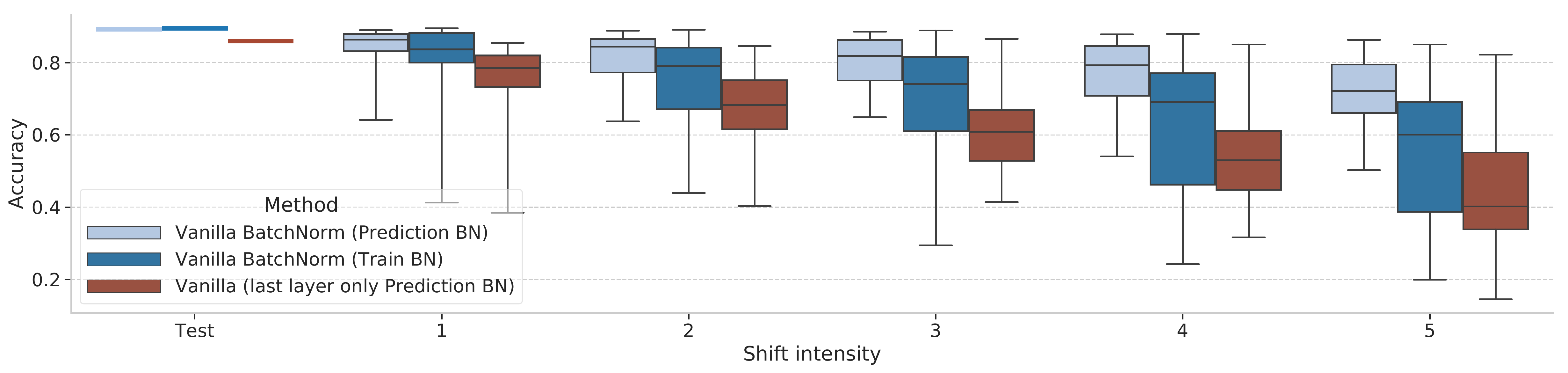}
      \includegraphics[width=0.95\linewidth]{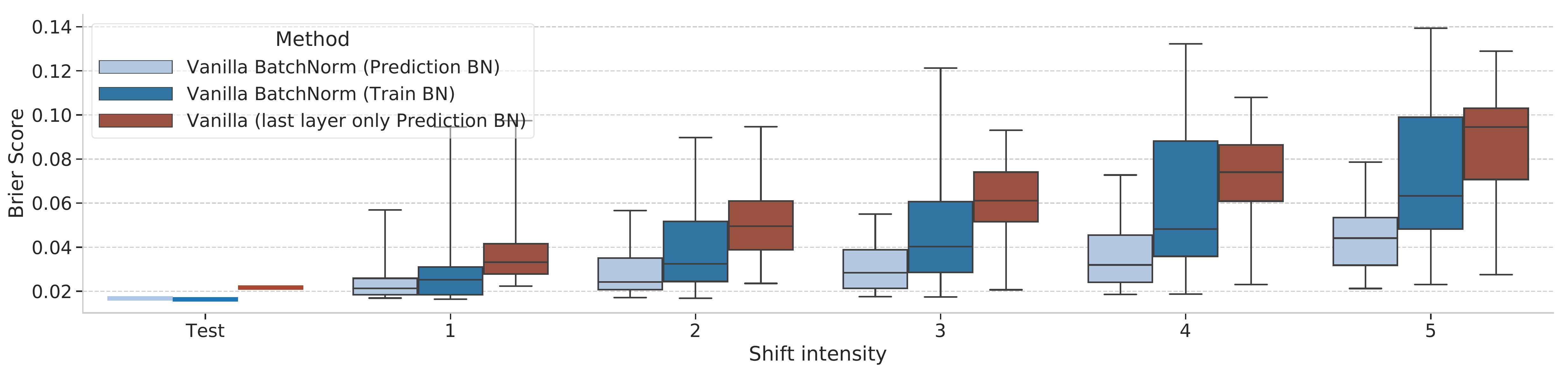}
    \caption{Calibration and accuracy under covariate shift with the CIFAR-10-C vanilla model compared to an altered Resnet-20 model where we have removed all Batch Norm layers and added one before the final linear layer.}
    \label{fig:cifar10_test_last_layer_only_all}
\end{figure}

\end{document}